\pdfoutput=1

\documentclass[11pt]{article}

\usepackage{acl}

\usepackage{times}
\usepackage{latexsym}

\usepackage[T1]{fontenc}

\usepackage[utf8]{inputenc}

\usepackage{microtype}
\usepackage{acronym}
\usepackage{amsmath}

\usepackage{times}
\usepackage{microtype}
\usepackage{graphicx}
\usepackage{subcaption}
\usepackage{cleveref}
\usepackage{booktabs}
\usepackage[inline]{enumitem}
\usepackage{amssymb}
\usepackage[skip=0pt]{caption}
\usepackage{booktabs}
\usepackage{makecell}
\usepackage{multirow}
\usepackage[ruled,vlined,linesnumbered]{algorithm2e}
\usepackage{makecell}

\acrodef{PDD}{prompt-then-distill debiasing}
\acrodef{NLP}{natural language processing}
\acrodef{DNN}{deep neural network}
\acrodef{OOD}{out-of-distribution}
\acrodef{ID}{in-distribution}

\acrodef{LLM}{large language model}
\acrodef{QA}{question answering}
\acrodef{CQA}{conversational question answering}
\acrodef{CQG}{conversational question generation}
\acrodef{NLI}{natural language inference}
\acrodef{PoE}{Product-of-Expert}
\acrodef{ICL}{in-context learning}
\acrodef{PT}{prompt tuning}
\acrodef{NLP}{natural language processing}
\acrodef{MOOD}{multi-objective optimization based debiasing}
\acrodef{ZOE}[SOD]{self-supervised position debiasing}
\acrodef{ICL}{in-context learning}
\acrodef{KGC}{knowledge-based conversation}
\acrodef{CE}{cross-entropy}
\acrodef{NLL}{negative log-likelihood}
\acrodef{CDA}{counterfactual data augmentation}
\acrodef{MSA}[OAM]{objective alignment}

\newcommand\model{\ac{ZOE}}

\definecolor{darkgreen}{RGB}{0,100,0}
\definecolor{forestgreen}{RGB}{34,139,34}

\allowdisplaybreaks
\usepackage{array}
\usepackage{tabularx}
\usepackage{svg}
\usepackage{colortbl}
\usepackage{float}
\usepackage{multicol}

\SetKwInput{KwInput}{Input} 
\SetKwInput{KwOutput}{Output}

\title{Self-Supervised Position Debiasing for Large Language Models}

\author{Zhongkun Liu\textsuperscript{\rm 1}, \ Zheng Chen\textsuperscript{\rm 1}, \ Mengqi Zhang\textsuperscript{\rm 1}, \ Zhaochun Ren\textsuperscript{\rm 2},  \ Zhumin Chen\textsuperscript{\rm 1}$^*$, \  Pengjie Ren\textsuperscript{\rm 1}\thanks{\ Corresponding authors.} \\
  \textsuperscript{\rm 1}School of Computer Science and Technology, Shandong University, China \\
  \textsuperscript{\rm 2} Leiden University, Leiden, the Netherlands \\
  \texttt{\{liuzhongkun,chenzheng01\}@mail.sdu.edu.cn}, \\
  \texttt{\{mengqi.zhang,chenzhumin,renpengjie\}@sdu.edu.cn},\\
\texttt{\ z.ren@liacs.leidenuniv.nl}}

\begin{document}
\maketitle

\begin{abstract}
Fine-tuning has been demonstrated to be an effective method to improve the domain performance of \acp{LLM}.
However, \acp{LLM} might fit the dataset bias and shortcuts for prediction, leading to poor generation performance.
Previous works have proven that \acp{LLM} are prone to exhibit position bias, i.e., leveraging information positioned at the beginning or end, or specific positional cues within the input.
Existing debiasing methods for \acp{LLM} require external bias knowledge or annotated non-biased samples, which is lacking for position debiasing and impractical in reality. 
In this work, we propose a \model~framework to mitigate position bias for \acp{LLM}.
\model~leverages unsupervised responses from pre-trained \acp{LLM} for debiasing without relying on any external knowledge.
To improve the quality of unsupervised responses, we propose an \ac{MSA} module to prune these responses.
Experiments on eight datasets and five tasks show that~\model~consistently outperforms existing methods in mitigating three types of position biases.
Besides, \model~achieves this by sacrificing only a small performance on biased samples, which is general and effective.
To facilitate the reproducibility of the results, we share the code of all methods and datasets on \url{https://github.com/LZKSKY/SOD}.
\end{abstract}
\acresetall

\acresetall
\section{Introduction}
Although \acp{LLM} have demonstrated remarkable unsupervised ability in various tasks~\citep{kojima2022large}, fine-tuning still overtakes it under the task-specific setting~\citep{ding2023parameter}.
However, fine-tuned \acp{LLM} might rely on the dataset biases and artifacts as shortcuts for prediction, as the fine-tuning datasets are sometimes skewed due to budget constraints~\citep{du2022shortcut}.
This results in poor generalization performance when applying fine-tuned \acp{LLM} to unseen test data and these models are vulnerable to various types of adversarial attacks~\citep{meade2022empirical}.

\begin{figure}[t]
    \centering
    \includegraphics[width=0.48\textwidth]{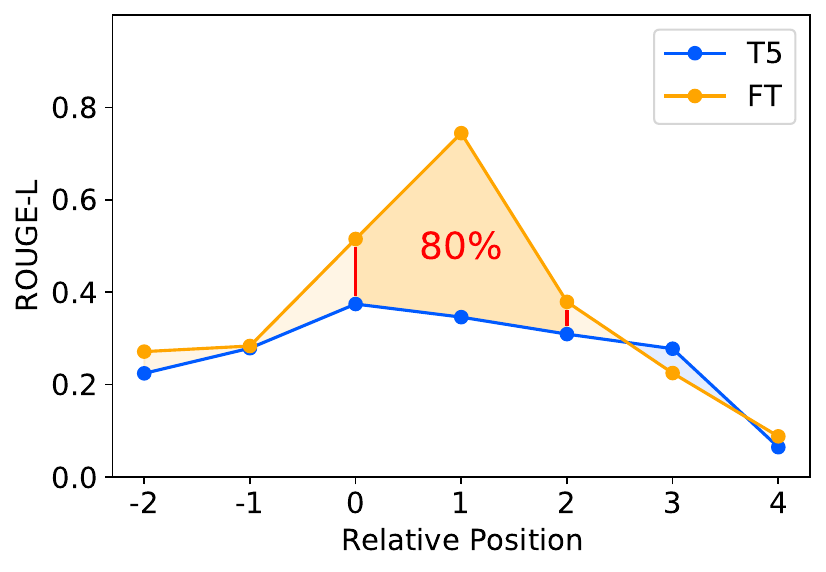}
    \caption{
    Question answering performance of FlanT5-large (T5) and fine-tuned FlanT5-large (FT) over different relative positions in CANARD. 
    Relative position means the distance of grounded utterances between the last turn answer and the current turn answer. 
    }
    \label{fig:intro_example}
\end{figure}

Position bias has been demonstrated to exist across various fine-tuned \acp{LLM}~\citep{liu2023lost}.
Specially, the well-known \acp{LLM}, e.g., GPT-3.5\footnote{\url{https://openai.com/blog/gpt-3-5-turbo-fine-tuning-and-api-updates}}, longchat-13B\footnote{\url{https://lmsys.org/blog/2023-06-29-longchat}}, are skilled when the relevant information occurs at the beginning or end of the input context, while the performance significantly degrades when \acp{LLM} need to find relevant information in the middle of the context. 
Analysis of \ac{CQA} on CANARD~\citep{CANARD} dataset further confirms the existence of position bias.
As shown in Fig.~\ref{fig:intro_example}, 80\% of the performance improvement after fine-tuning is attributed to fitting bias on relative positions 0-2.
This encourages researchers to engage in position debiasing~\cite{meade2022empirical}.

Early works mainly focus on mitigating position bias on extractive tasks before the emergence of \acp{LLM}~\citep{ko2020look,karimi2020end}.
A prominent debiasing method is \ac{PoE}, which discourages the extractive model from learning position bias picked up by the fixed biased model~\citep{du2021towards,shinoda2022look}.
Recently, more and more works focus on debiasing for generative models, e.g., \acp{LLM}~\citep{guo2022auto,li2023prompt}.
They adopt either \ac{ICL} to guide the generation of \ac{LLM}~\citep{meade2023using} or \ac{PT} to fine-tune prompts for \acp{LLM}~\citep{li2023prompt}.
However, these works mostly focus on mitigating social bias~\citep{kasneci2023chatgpt}, e.g., gender bias and racial bias, leaving position debiasing unexplored.
Besides, these works cannot be transferred to mitigate position bias, as they require manually annotated non-biased samples for \ac{ICL} or external bias knowledge for \ac{PT}, which are lacking for position debiasing.

To deal with this challenge, we propose to leverage the low position bias characteristics of pre-trained \acp{LLM}.
Previous studies have shown that pre-trained \acp{LLM} are more robust to position bias~\citep{UtamaMSG21}.
This is due to the randomness of knowledge utilization in generation during pre-training.

As shown in Fig. \ref{fig:intro_example}, the ROUGE-L score of pre-trained T5 fluctuates within the range of 0.2 to 0.4 across almost all relative positions, demonstrating its robustness against position bias.

In this paper, we propose a \model~framework to mitigate position bias for \acp{LLM}.
First, we use a low-bias inference module to collect unsupervised responses with low position bias by applying various prompting strategies.
Then, we propose an \ac{MSA} module to prune the unsupervised responses, as low-quality responses will undermine model performance on non-biased samples.
Finally, we use a multi-objective optimization module to leverage these unsupervised responses for fine-tuning.
The whole process does not require any external bias knowledge or non-biased samples, which is self-supervised and general.

To verify the effectiveness of \model, we conduct experiments on eight datasets covering five tasks.
Experimental results show that the \model~achieves superior performance in mitigating three types of position bias significantly, including lead bias, relative position bias, and lexical bias.
The main contributions of this work are as follows.
\begin{itemize}[leftmargin=*, itemsep=0pt, topsep=5pt]
    \item We propose to mitigate position bias for \acp{LLM} in a self-supervised setting, i.e., without any external knowledge or annotated samples.
    \item We propose a \acs{ZOE} framework for position debiasing with an \acs{MSA} module to prune low-quality unsupervised responses for fine-tuning.
    \item Experiments show that \model~can mitigate various types of position biases by sacrificing only small performance on biased samples, demonstrating its effectiveness and generality.
\end{itemize}

\section{Preliminary}
\subsection{Task Definition}
Given a biased dataset $D$ for the target task, our position debiasing task aims to improve the model robustness against position bias when fine-tuning on target tasks, i.e., to achieve superior performance on non-biased samples by retaining the performance on biased samples.
Here, the biased samples exhibit similar position bias as training samples, and the non-biased samples do not contain these position clues.
The target task for fine-tuning can be any \ac{NLP} tasks exhibiting position bias. 
In this paper, we focus on five target tasks: \acf{CQA}, \acf{CQG}, \acf{KGC}, summarization and \acf{NLI}.

\begin{figure*}[th]
    \centering
    \includegraphics[width=\textwidth]{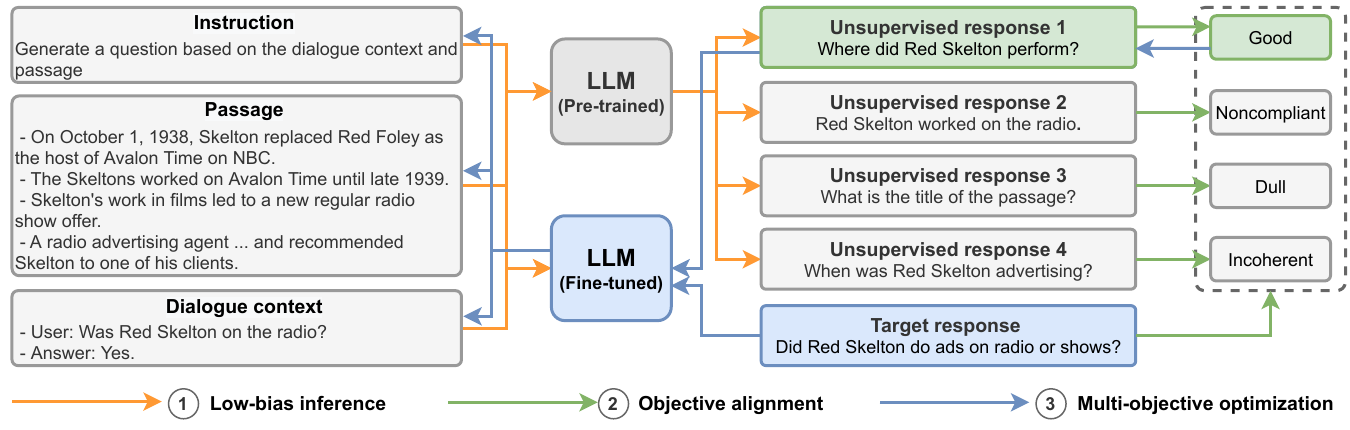}
    \caption{
    Overview of our proposed \acf{ZOE} framework (taking \ac{CQG} as the example).
    First, the low-bias inference module collects multiple unsupervised questions from \acp{LLM}.
    Then, the objective alignment module aligns these questions with the target question.
    Finally, these aligned questions are utilized for fine-tuning within the multi-objective optimization module.
    }
    \label{fig:model}
\end{figure*}

\subsection{Large Language Model}
\Acp{LLM} have attracted much attention and become state-of-the-art due to their remarkable ability of language generation.
They formulate all \ac{NLP} tasks as language generation tasks with different task prompts:
\begin{equation}
\label{eq:lm_gen}
\begin{split}
    p(y) &= p(y | prompt, x) \\
         &= \prod_t p(y_{t+1} | prompt, x, y_1, y_2, ..., y_t),
\end{split}
\end{equation}
where $x$, $y$ and $prompt$ are task input, output, and task prompt, respectively.
$y_i$ denotes the $i$-th token in $y$.
Task prompt $prompt$ consists of the task instruction and demonstrations to tell the \acp{LLM} the definition of the task and how it works.
The introduction of task prompt enables \acp{LLM} to utilize all available data for training and improve their generalization ability on unseen tasks.

\section{Method}
We propose a \acf{ZOE} framework to mitigate position bias for generative \acp{LLM}.
As shown in Fig.~\ref{fig:model}, \model~consists of three modules: low-bias inference, \acf{MSA}, and multi-objective optimization, where all modules do not require external bias knowledge or non-biased samples.
The Low-bias inference module generates unsupervised responses with lower position bias by utilizing pre-trained \acp{LLM} (in \S \ref{sec:inference}).
Subsequently, the \ac{MSA} module is employed to prune these low-quality unsupervised responses based on the target responses (in \S \ref{sec:filtering}).
Finally, the multi-objective optimization module fine-tunes the \acp{LLM} by optimizing the task objective and the debiasing objective (in \S \ref{sec:finetune}).
The task objective utilizes target responses to enhance task-specific performance, and the debiasing objective leverages unsupervised responses for position debiasing.

\subsection{Low-bias Inference} \label{sec:inference}
Given a biased training dataset $D=\{(x_i, y_i)\}_{i=1}^N$ for the target task, the low-bias inference module generates unsupervised responses with low position bias based on pre-trained \acp{LLM}~\cite{UtamaMSG21}.

We employ three prompting strategies for generation and adapt them for different target tasks.
\begin{itemize}[leftmargin=*,nosep]
    \item \textbf{Instruction-only prompting} generates respon-\\ses of target task by feeding the task input and task instruction to the pre-trained \acp{LLM}.
    Concretely, we assign the $prompt$ by task instruction in Eq.~\ref{eq:lm_gen} for the generation.

    \item \textbf{Diverse prompting} generates responses with diverse aspects by feeding various prompts to \acp{LLM}.

    \item \textbf{\Acf{ICL}} also feeds multiple input-output examples to \acp{LLM} for generation, in addition to the task instruction and input.
    It enhances the model comprehension of target tasks but requires a longer input length.

\end{itemize}
We adopt \ac{ICL} only for \ac{NLI}, due to the limit of the model input length.
We employ diverse prompting for \ac{CQG}, which is intrinsically creative and diverse.
And instruction-only prompting is implemented for \ac{CQA}, \ac{KGC}, and summarization.

\subsection{Objective Alignment} \label{sec:filtering}
Unlike the annotated high-quality target responses for the task objective, the unsupervised responses for the debiasing objective are of lower quality and noisy, because they are generated by pre-trained models without specific fine-tuning on the target task. 
To reduce the interference between the debiasing objective and the task objective, we propose an \acf{MSA} module to prune the unsupervised responses $y'_i$ to better align with the target response $y_i$. 

We propose various alignment strategies for different tasks based on their intrinsic characteristics.

\textbf{Alignment for tasks excluding \acs{NLI}.}
We first identify low-quality unsupervised responses and then drop them, as the generated responses are flexible for modification and modification may introduce new errors.
We apply and combine four strategies to identify them.
\begin{itemize}[leftmargin=*,nosep]
    \item \textbf{Non-compliant identification} identifies unsupervised responses deviating from the task instruction by keyword matching.
    For example, it identifies non-`what' questions when `what' is specified in the instruction.

    \item \textbf{Dull identification} identifies dull responses by keyword matching, e.g., ``What is the title of the passage?''.
    \\
    \item \textbf{Incoherent identification} identifies incoherent responses if the perplexity of any token in the response falls below a pre-defined threshold.

    \item \textbf{Unreliable identification} identifies unreliable responses if the overlap score between unsupervised and target responses is less than a pre-defined threshold.
    The intuition is that the fact in the response may be wrong if its semantics deviate significantly from the fact in the reference.

\end{itemize}

For \ac{CQA}, summarization, and \ac{KGC} tasks, we align unreliable responses since the facts in answers, summaries, and knowledge-enriched responses are always unique in their semantics.
For \ac{CQG} task, we align non-compliant, dull, or incoherent responses, considering that the appropriate questions are diverse in semantics.

\textbf{Alignment for \acs{NLI}.}
As the generated responses for \ac{NLI} are deterministic, i.e., entailment, neutral, and contradiction, we can directly estimate the probability distribution over all classes by prompting and then align it.
The estimated probability distributions are low-quality sometimes when the target class dominates, which is redundant for optimizing the task objective and strengthens position bias.
Therefore, we align the estimated probability distribution by masking the target class: 
\begin{equation}
\label{eq:nli_y}
    y'_i = s_i \cdot mask_i,
\end{equation}
where $mask_i$ is a vector to mask the target class:
\begin{equation}
\label{eq:nli_mask}
    mask_{i,j} = 
        \begin{cases}
        0, \text{ if } class_j = y_i \\
        1, \text{otherwise}.
        \end{cases}
\end{equation}
Here, $class_j$ is the tokens for $j$-th class and $y_i$ is the tokens for the target class.
$s_i$ is the probabilities distribution over all classes in \ac{NLI} inference:
\begin{equation}
\label{eq:nli_gen}
\begin{split}
         s_i &= [s_{i, 1}, s_{i, 2}, ..., s_{i, |class|}] \\
    s_{i, j} &= \frac{p(class_j|x_i)}{\sum_j^{|class|}p(class_j|x_i)},
\end{split}
\end{equation}
where $p(class_j|x_i)$ is the generation probabilities of $j$-th class tokens, which is calculated by Eq.~\ref{eq:lm_gen}.

\subsection{Multi-Objective Optimization} \label{sec:finetune}
Given the target response $y_i$, aligned unsupervised response $y'_i$, and input $x_i$, our multi-objective optimization module fine-tunes the model to generate task-specific but low-bias responses.
It fine-tunes the model by optimizing two objectives: target responses as task objective to improve the performance on the target task and unsupervised responses as debiasing objective to mitigate position bias:
\begin{equation}
\begin{split}
    \mathcal{L} =& (1 - \alpha) \cdot \mathcal{L}_{target}(x_i, y_i) + \\
    & \alpha \cdot \mathcal{L}_{align}(x_i, y'_i),
\end{split}
\end{equation}
where $\alpha$ is a hyper-parameter for tradeoff and $\mathcal{L}_{target}(x_i, y_i)$ is a \ac{NLL} loss to maximize the generation probability of target response:
\begin{equation}
    \mathcal{L}_{target}(x_i, y_i) = -\sum_{j=1}^{|y_i|}\log p(y_{i,j}|x_i, y_{i, <j}).
\end{equation}

$\mathcal{L}_{align}(x_i, y'_i)$ is a task-specific objective to mitigate position bias.

\textbf{For tasks excluding \acs{NLI},} such as \ac{CQA} and \ac{CQG}, summarization and \ac{KGC}, we use \ac{NLL} loss to maximize the probability of generating $y'_i$:
\begin{equation}
    \mathcal{L}_{align}(x_i, y'_i) = -\sum_{j=1}^{|y'_i|}\log p(y'_{i,j}|x_i, y'_{i, <j}).
\end{equation}

\textbf{For \acs{NLI},} we maximize the generation probability of the most likely class tokens after alignment:
\begin{equation}
    \mathcal{L}_{align}(x_i, y'_i) = -s_{i,ind(i)}\log p(class_{ind(i)}|x_i),
\end{equation}
where ${ind(i)}$ is the index of the class, $class_{ind(i)}$ is the class tokens, and $s_{i,ind(i)}$ is the generation probability of the class tokens in Eq.~\ref{eq:nli_gen}:
\begin{equation}
    ind(i) = \arg \max y'_i.
\end{equation}
$y'_i$ is the masked probability distribution for all classes in Eq.~\ref{eq:nli_y}.

\section{Experiments}
We evaluate \model~on three categories of \ac{NLP} tasks based on changes in conveyed information from input to output: language understanding tasks, language compression tasks and language creation tasks~\cite{deng2021compression}.
Language understanding tasks (e.g., \ac{NLI}, \ac{CQA}) aim to comprehend and interpret natural language input given a conversation or document context.
For a compression task (e.g., summarization), the goal is to concisely describe the most important information in the input (e.g., a document).
A creation task (e.g., \ac{CQG}, \ac{KGC}) generates output that adds new information on top of input (e.g., dialogue history).

\subsection{Datasets}
We conduct experiments on eight widely used benchmark datasets: CANARD~\citep{CANARD}, CoQAR~\citep{brabant2022coqar}, CNN/DM~\citep{nallapati2016abstractive}, Newsroom~\citep{grusky2018newsroom}, Doc2dial~\citep{feng2020doc2dial}, Mutual~\citep{cui2020mutual}, SNLI~\citep{bowman2015large} and QNLI~\citep{wang2018glue}, covering five \ac{NLP} tasks: \ac{CQA}, \ac{CQG}, summarization, \ac{KGC} and \ac{NLI}.
Following previous works~\citep{ko2020look,shinoda2022look}, we split the test dataset into biased dataset and non-biased dataset for simulation depending on the bias type in each dataset.
The details of datasets and dataset splitting are provided in \S \ref{sec: app_datasets} and \S \ref{sec: app_dataset_split}.

\subsection{Evaluation Metrics}
Following previous works~\citep{chen2019evaluating,nallapati2016abstractive,tuan2020capturing,meng2020dukenet}, we adopt ROUGE-L~\cite{lin2004rouge} as evaluation metrics for \ac{CQA}, \ac{CQG}, summarization and \ac{KGC} tasks, in which ROUGE-L has been shown to correlate well with human evaluation~\citep{liu2008correlation}.
We use macro-accuracy for the classification task, \ac{NLI}. 
We use nlg-eval package\footnote{\url{https://github.com/Maluuba/nlg-eval}} for the implementation of evaluation metrics.

\subsection{Baseline Methods}

\begin{itemize}[leftmargin=*,nosep]
\item \textbf{BASE} is the pre-trained LLM with unsupervised instruction-following fine-tuning.

\item \textbf{Random Position~(RP)}~\citep{shinoda2022look} randomly perturbs input positions to reduce the model's dependence on token positions in prediction.

\item \textbf{Fine-tune~(FT)} is the \ac{LLM} fine-tuned on the dataset for the target task to improve the performance of the target task.

\item \textbf{MarCQAp}~\citep{gekhman2023robustness} is a novel prompt-based history modeling approach for \ac{CQA} and \ac{CQG} that highlights answers from previous conversation turns by inserting textual prompts in their respective positions.

\item \textbf{Minimax}~\citep{korakakis2023improving} is an \ac{NLI} model which leverages an auxiliary model to maximize the loss of the \ac{NLI} model by up-weighting `hard' samples, thus reducing its reliance of shortcuts in `easy' samples.

\item \textbf{GenX}~\citep{varab2023abstractive} is a new summarization paradigm that unifies extractive and abstractive summarization with generative modeling.

\item \textbf{SG-CQG}~\citep{do2023modeling} is a state-of-the-art \ac{CQG} models with two stages: what-to-ask for rational span selection in the referential document and how-to-ask for question generation.

\item \textbf{FocusL}~\citep{deng2023towards} is a debiasing method built for \ac{KGC} by adaptively re-weighting the loss of each token, thus encouraging the model to pay special attention to knowledge utilization.

\end{itemize}

\subsection{Implementation Details}
We use FlanT5-large~\citep{chung2022scaling} as the base \acs{LLM} for all models.
The hidden size is 768.
We use the Adam optimizer with a default learning rate $1e^{-4}$~\citep{KingmaB14} and set gradient clipping with a default maximum gradient norm of 1.0\footnote{\url{https://huggingface.co/docs/transformers/v4.33.0/en/main_classes/trainer}}.
We select the best model based on the BLEU@2 or macro-accuracy score on the validation set.
We use $\alpha$=0.2 for \ac{CQA} on CoQAR, \ac{NLI} and \ac{KGC} tasks and $\alpha$=0.1 for other tasks, by default.
We set the pre-defined thresholds for incoherent identification and unreliable identification from 0.1, 0.15 and 0.2 and select the one that maintains approximately 20\% unsupervised responses.
We run all experiments with NVIDIA RTX3090 24 GB GPU cards.

\section{Results}

\begin{table*}[ht]
\centering
\caption{Overall performance (\%) on language understanding tasks. Boldface indicates the best results in terms of the corresponding dataset.}
\label{tab: lu_task} 
\setlength{\tabcolsep}{4pt}
\begin{tabular}{l c c c c c c c c c}
\toprule
\multirow{3}{*}{Method} & \multicolumn{4}{c}{NLI~(\%)} & \multicolumn{4}{c}{CQA~(\%)} \\
 & \multicolumn{2}{c}{SNLI} & \multicolumn{2}{c}{QNLI} & \multicolumn{2}{c}{CoQAR} & \multicolumn{2}{c}{CANARD} \\ 
\cmidrule(lr){2-3}\cmidrule(lr){4-5}\cmidrule(lr){6-7}\cmidrule(lr){8-9}
 & Biased & Non-Biased & Biased & Non-Biased & Biased & Non-Biased & Biased & Non-Biased \\
\midrule
BASE & 77.5 & 79.0 & 90.3 & 88.3 & 47.4 & 39.8 & 34.0 & 17.3 \\
RP & --- & --- & --- & --- & 61.5 & 51.6 & 60.4 & 20.4 \\
MarCQAp & --- & --- & --- & --- & \textbf{66.9} & 52.3 & 66.3 & 21.3 \\
Minimax & 91.6 & 87.0 & 90.9 & 89.7 & --- & --- & --- & --- \\
FT & \textbf{92.0} & 88.0 & \textbf{94.3} & 89.8 & 64.6 & 51.9 & \textbf{67.5} & 20.8 \\
\rowcolor{gray!20}
\acs{ZOE} & \textbf{92.0} & \textbf{88.4} & \textbf{94.3} & \textbf{90.9} & 66.3 & \textbf{53.7} & 65.7 & \textbf{21.9} \\
\bottomrule
\end{tabular}

\end{table*}

\begin{table*}[ht]
\centering
\caption{Overall performance (\%) on language creation tasks. 
Boldface indicates the best results in terms of the corresponding dataset.}
\label{tab: lc_task} 

\setlength{\tabcolsep}{4pt}
\begin{tabular}{l c c c c c c c c c}
\toprule
\multirow{3}{*}{Method}  & \multicolumn{4}{c}{CQG~(\%)} & \multicolumn{4}{c}{KGC~(\%)} \\
 & \multicolumn{2}{c}{CoQAR} & \multicolumn{2}{c}{CANARD} & \multicolumn{2}{c}{Doc2dial} & \multicolumn{2}{c}{Mutual} \\ 
\cmidrule(lr){2-3}\cmidrule(lr){4-5}\cmidrule(lr){6-7}\cmidrule(lr){8-9}
 & Biased & Non-Biased & Biased & Non-Biased & Biased & Non-Biased & Biased & Non-Biased \\
 \midrule
BASE & 16.0 & 17.0 & 17.9 & 17.1 & 23.9 & 13.7 & 25.4 & 21.3 \\
RP & 23.2 & 18.3 & 24.6 & 21.2 & 35.3 & 32.2 & 89.6 & 46.1 \\
MarCQAp & 19.7 & 17.5 & \textbf{26.0} & 21.8 & --- & --- & --- & --- \\
SG-CQG & 14.9 & 15.5 & 19.7 & 17.8 & --- & --- & --- & --- \\
FocusL & --- & --- & --- & --- & 39.8 & 35.9 & 83.0 & 51.7 \\
FT & \textbf{26.7} & 17.2 & \textbf{26.0} & 21.6 & 38.7 & 36.0 & 93.9 & 38.9 \\
\rowcolor{gray!20}
\acs{ZOE} & \textbf{26.7} & \textbf{18.8} & 25.9 & \textbf{22.4} & \textbf{40.6} & \textbf{38.3} & \textbf{94.0} & \textbf{53.0}     \\ 
\bottomrule
\end{tabular}
\end{table*}

\begin{table}[ht]
\centering
\caption{Overall performance (\%) on language compression tasks. 
Boldface indicates the best results in terms of the corresponding dataset.}
\label{tab: lcomp_task} 

\setlength{\tabcolsep}{2pt}
\begin{tabular}{l c c c c c}
\toprule
\multirow{3}{*}{Method} & \multicolumn{4}{c}{Summarization~(\%)} \\
 & \multicolumn{2}{c}{CNN/DM} & \multicolumn{2}{c}{Newsroom} \\ 
\cmidrule(lr){2-3}\cmidrule(lr){4-5}
 & Biased & Non-Biased & Biased & Non-Biased \\
 \midrule
BASE & 22.3 & 11.6 & 35.1 & 20.0 \\
RP & 23.9 & 15.1 & 47.5 & \textbf{22.0} \\
GenX & 17.6 & 13.7 & 29.2 & 19.0 \\
FT & 26.9 & 16.8 & \textbf{51.1} & 19.8 \\
\rowcolor{gray!20}
\acs{ZOE} & \textbf{27.1} & \textbf{17.3} & 50.9 & 21.3 \\
\bottomrule
\end{tabular}

\vspace{-3.5mm}
\end{table}

The overall performances of all methods on language understanding tasks, language creation tasks and language compression tasks are listed in Table~\ref{tab: lu_task}-\ref{tab: lcomp_task}.
We have three main observations from the results.

First, \acp{LLM} are susceptible to the bias in the dataset after fine-tuning.
As we can see in Table~\ref{tab: lu_task}, FT achieves 34.7\% improvement on the biased dataset of CoQAR, but 8.6\% improvement on the non-biased dataset.
This is because \acp{LLM} can easily overfit the shortcut of the training dataset in fine-tuning, just like existing neural networks~\citep{ko2020look}.

Second, \model~can mitigate position bias significantly on almost all datasets for three types of tasks.
As shown in Table~\ref{tab: lu_task}-\ref{tab: lcomp_task}, \model~improves the performance on the non-biased dataset by 1\% to 2\% on almost all tasks, compared to FT and all baselines.
The reason is that \model~can leverage unsupervised responses with low position bias for optimization in multi-objective optimization module.

Third, \model~only sacrifices a small performance on the biased dataset when mitigating position bias.
As shown in Table~\ref{tab: lcomp_task}, RP achieves comparable performance to \model~on the non-biased dataset of Newsroom. 
However, the ROUGE-L of RP drops 3.6\% compared to FT on the biased dataset, while that of \model~only drops 0.2\%. 
This is because the perturbation in RP impairs the overall data quality for fine-tuning, while unsupervised responses in \model~are aligned to improve the quality in \S \ref{sec:filtering}.

Note that some baselines achieve poor performance, sometimes lower than BASE.
First, in Table~\ref{tab: lcomp_task}, GenX performs worse even than BASE on the biased dataset of Newsroom.
The reason is that the summarization datasets are abstractive and suitable for generative models, e.g., T5, while GenX is an extractive baseline.
Second, in Table~\ref{tab: lc_task}, the performance of SG-CQG is worse than that of FT on the biased dataset of CoQAR.
This is because SG-CQG is a three-stage question generation model focusing on improving question diversity.
The selected answer span for question generation is randomly chosen from massive generated candidates.

Besides, to further verify the effectiveness of \model, we have done t-test significant tests.
We found that \model~can achieve significant improvement over MarCQAp on CoQAR (\acs{CQA} and \acs{CQG}) and CANARD (\acs{CQG}), over FocusL on Doc2dial and Mutual, over FT on CNN/DM.
The above results indicate that \model~can mitigate position bias in most cases.

\section{Analysis}

In this section, we analyze the effect of the quality of unsupervised responses in \S \ref{sec: unsupervised_response} and objective weighting in the multi-objective optimization module in \S \ref{sec: objective_weighting}.
The overall results of all tasks are presented in \S \ref{sec: overall_analysis}.
Besides, we also conduct a case study in \S \ref{sec: case_study} and provide cases for all datasets in \S \ref{sec: cases}.

\subsection{Analysis of \acs{MSA}} \label{sec: unsupervised_response}

To analyze the effect of the \ac{MSA} module, we conduct analyses with unsupervised responses obtained from various sources with different qualities in Table~\ref{tab: response quality}.
\acs{ZOE} w/o \acs{MSA}, \acs{ZOE} w/ T5-base and \acs{ZOE} w/ T5-xlarge denote \model~using unsupervised responses without alignment, responses from FlanT5-base and FlanT5-xlarge, respectively.
We have two observations.

First, low-quality responses without \acs{MSA} leads to worse performance.
As we can see, \acs{ZOE} outperforms \acs{ZOE} w/o \acs{MSA} on non-biased datasets by leveraging \acs{MSA} for enhancing the response quality.
Poor-quality responses will undermine the model comprehension of the task, thus leading to worse performance. 

\begin{table}[t]
\centering
\caption{\model~performance (\%) of \ac{CQG} task using various unsupervised responses. 
`N-Biased' denotes performance on non-biased datasets.
Boldface indicates the best results in terms of the corresponding dataset.}
\label{tab: response quality} 

\setlength{\tabcolsep}{2pt}
\begin{tabular}{l c c c c c}
\toprule
\multirow{2}{*}{Method}  & \multicolumn{2}{c}{CoQAR} & \multicolumn{2}{c}{CANARD} \\ 
\cmidrule(lr){2-3}\cmidrule(lr){4-5}
 & Biased & N-Biased & Biased & N-Biased \\
 \midrule
\acs{ZOE} & \textbf{26.7} & \textbf{18.8} & 25.9 & \textbf{22.4} \\
- w/o \acs{MSA} & 25.8 & 18.0 & \textbf{26.1} & 21.5 \\
- w/ T5-base & 26.3 & 18.3 & \textbf{26.1} & 22.0 \\
- w/ T5-xlarge & 26.6 & 17.9 & 25.7 & 22.0 \\
FT & \textbf{26.7} & 17.2 & 26.0 & 21.6 \\
\bottomrule
\end{tabular}
\end{table}

Second, \acs{MSA} is robust to various sources of unsupervised responses.
\acs{ZOE} w/ T5-base and \acs{ZOE} w/ T5-xlarge perform worse than \acs{ZOE} on CANARD, but still outperform FT.
We infer that responses from other \acp{LLM} use different knowledge/parameters for generation, which mismatch with that of T5.
The difference amplifies the divergence between the task objective and the debiasing objective in \S \ref{sec:filtering}, thus leading to worse performance.
Even that, \acs{MSA} can reduce this divergence to better achieve both objectives.

\subsection{Analysis of Objective Weighting} \label{sec: objective_weighting}
\begin{figure}[ht]
    \centering
    
    \subfloat[CoQAR~(CQA)]{\vspace{-5pt}\includegraphics[width=0.23\textwidth]{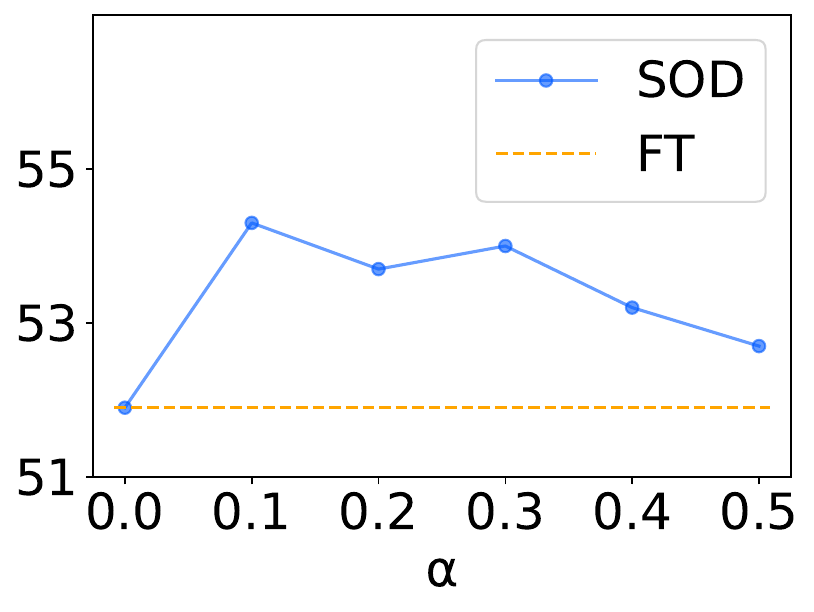}}
    \hfill
    \subfloat[CANARD~(CQG)]{\vspace{-5pt}\includegraphics[width=0.23\textwidth]{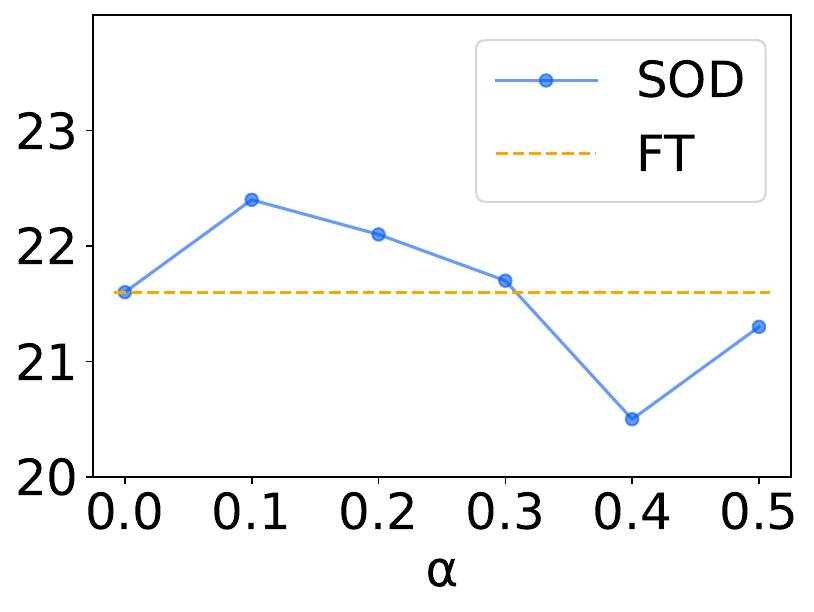}}
    \newline
    \subfloat[Mutual]{\vspace{-5pt}\includegraphics[width=0.23\textwidth]{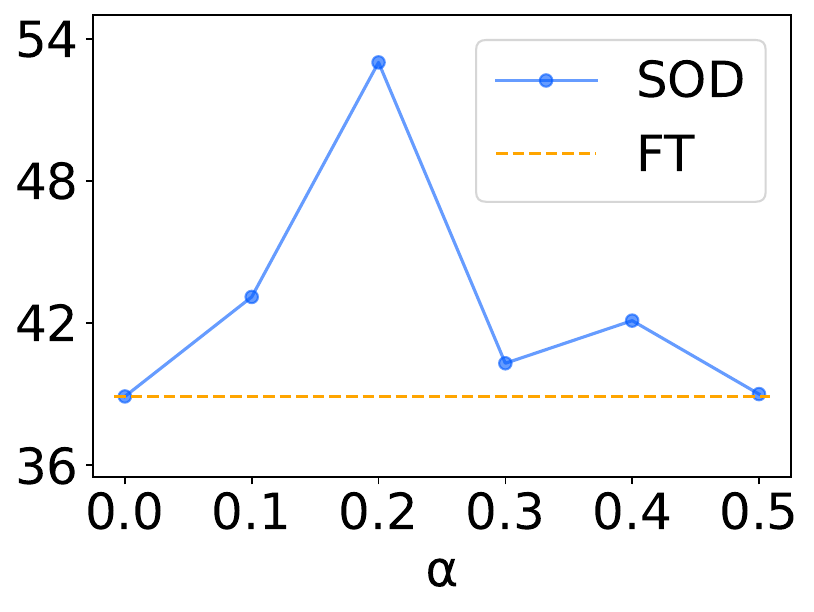}}
    \hfill
    \subfloat[QNLI]{\vspace{-5pt}\includegraphics[width=0.23\textwidth]{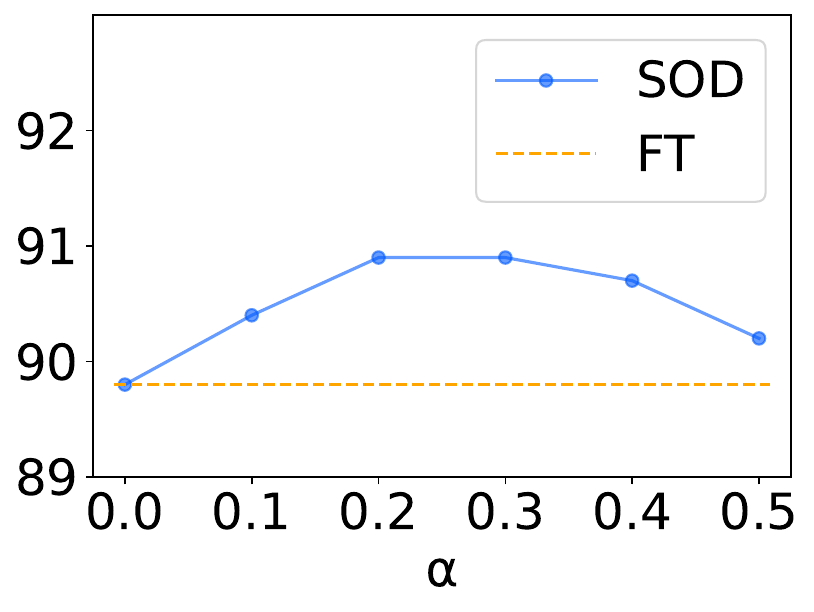}}

    \caption{Performance~(\%) of four tasks over each $\alpha$.
    The x-axis denotes the value of $\alpha$ and the y-axis denotes the ROUGE-L score on non-biased datasets.}
    \label{fig:range_alpha}
\vspace{-2mm}
\end{figure}

To analyze the effect of weighting on the debiasing objective, we present the performance of \model~using different $\alpha$ in Fig.~\ref{fig:range_alpha}.
We have two observations from the results.

First, the performance of \model~drops with the increase of the weight of unsupervised responses in multi-objective optimization.
In \ac{CQA} on CoQAR, the ROUGE-L score of \model~drops from 53.6\% to 52.7\% when increasing $\alpha$ from 0.1 to 0.5.
This is because the model performance depends on not only the degree of bias but also the data quality.
Increasing $\alpha$ will reduce the position bias of all responses, yet it will hurt the quality concurrently.

Second, our proposed \model~always outperforms FT under various $\alpha$.
As shown in Fig.~\ref{fig:range_alpha}, \model~performances of \ac{CQA} on CoQAR all exceed 52.5\% using various $\alpha$, while FT only achieves 52.0\%.
This demonstrates the effectiveness and robustness of \model~in mitigating position bias.

\subsection{Analysis on Training Samples}

\begin{figure}[ht]
    \centering
    \subfloat[CoQAR~(CQA)]{\vspace{-5pt}\includegraphics[width=0.23\textwidth]{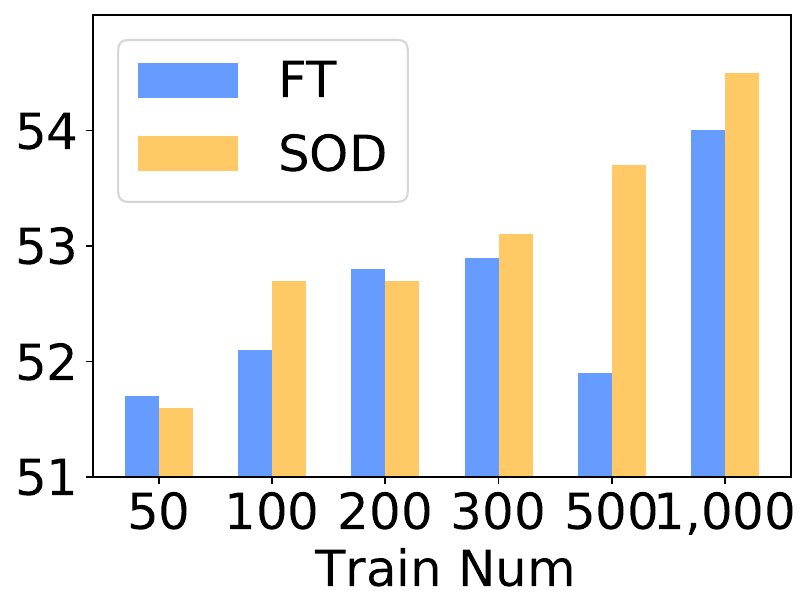}} 
    \hfill
    \subfloat[CANARD~(CQG)]{\vspace{-5pt}\includegraphics[width=0.23\textwidth]{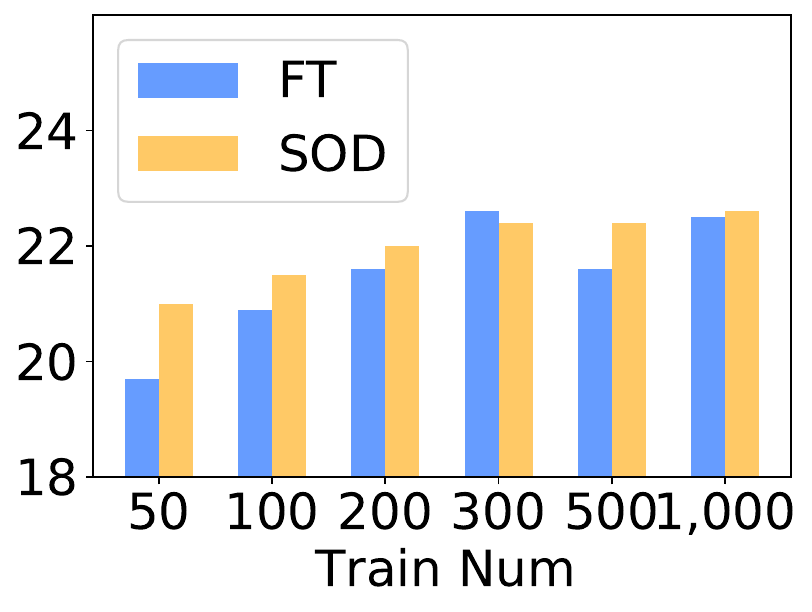}}
    \newline
    \subfloat[Mutual]{\vspace{-5pt}\includegraphics[width=0.23\textwidth]{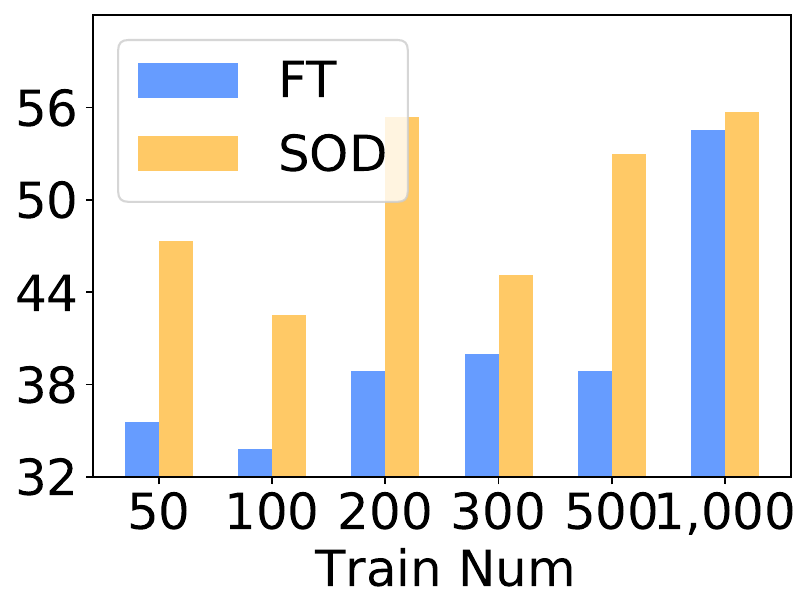}}
    \hfill
    \subfloat[QNLI]{\vspace{-5pt}\includegraphics[width=0.23\textwidth]{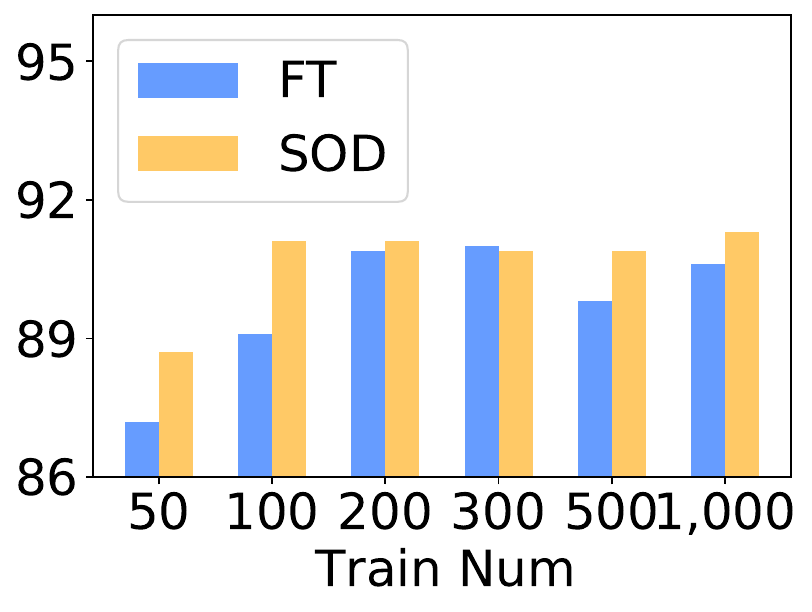}} 
    \caption{Performance~(\%) of four tasks over different numbers of training samples. 
    The x-axis denotes the number of training samples and the y-axis denotes the ROUGE-L score on non-biased datasets.}
    \label{fig:range_num}
\end{figure}

We also analyze the effect of the number of training samples to verify the effectiveness of \model~under various low-resource settings.
We plot the results in Fig.~\ref{fig:range_num} and have two observations.

First, nearly all methods perform better when increasing training samples.
As we can see in Fig.~\ref{fig:range_num}, the ROUGE-L score of FT increases from 19.7\% to 22.5\% on CANARD when the number of training samples increases from 50 to 1,000.
Increasing training samples can improve the generalization ability of \acp{LLM}, thus leading to better performance.

Second, our proposed \model~outperforms FT under various low-resource settings.
As shown in Fig.~\ref{fig:range_num}, the ROUGE-L scores of \acs{ZOE} depicted by the orange bars are consistently higher than those of FT in blue.
This is because there are always around 40\% aligned unsupervised responses for fine-tuning regardless of the variation in the number of training samples, which are enough for effective debiasing.

\section{Related Work}
\subsection{Position Bias and Debiasing}
Position bias has been explored mainly in three extractive tasks, i.e., \ac{NLI}, summarization and \ac{CQA}.
In, \ac{NLI}, \citet{gururangan2018annotation,poliak2018hypothesis} show that the class labels are highly correlated to certain words in the hypothesis.
\citet{mccoy2019right} report that models always rely on word overlap between hypothesis and premise for prediction.
\citet{karimi2020end,du2021towards} train a robust \ac{NLI} model in an ensemble manner by \ac{PoE}.
They train a hypothesis model to learn the lexical bias, which guides the \ac{NLI} model to focus on other patterns in the dataset that generalize better.
\citet{ghaddar2021end} re-weight the importance of easy and hard samples to prevent the model from fitting the bias.
As the weights are derived from the model itself, this method does not require any external annotations.

Similar findings are reported in \ac{CQA} tasks.
\citet{weissenborn2017making,sugawara2018makes} demonstrate that only using partial inputs is sufficient to correctly extract the answer span for the question in most cases.
\citet{ko2020look} address that absolute position of answer spans can work as a spurious clue for prediction.
\citet{shinoda2022look} report that relative position of answer spans is another clue for position bias.
To mitigate position bias, \citet{ko2020look,shinoda2022look} build bias ensemble models by \ac{PoE} similar to \citet{karimi2020end}. 
They design biased models with position-only features to guide \ac{CQA} models to rely more on semantic features for answering.

\citet{kedzie2018content,grenander2019countering} find that 58\% of selected summary utterances come directly from the lead utterances, and models trained on these articles perform considerably worse when utterances in the article are randomly shuffled.
To mitigate lead bias, a simple but effective method is to randomly shuffle the document for training~\citep{grenander2019countering}.
Then, \citet{xing2021demoting} uses adversarial training for debiasing.
They design a position prediction module and optimize the reverse loss for position prediction, forcing the encoder to leverage non-position features.

However, existing works on mitigating position bias always focus on extractive tasks.
They can hardly be transferred to generative tasks as the label space in generative tasks is too large for bias estimation or adversarial training.
In this paper, we focus on position debiasing for generative \acp{LLM}.

\subsection{Debiasing for \acp{LLM}}
Works on debiasing for \acp{LLM} always focus on social bias, e.g., gender bias and racial bias, rather than position bias~\citep{meade2022empirical,du2022shortcut}.
Existing works on mitigating social bias for \acp{LLM} can be classified into three types: pre-processing methods, in-processing methods and post-processing methods.

In pre-processing, \citet{zmigrod2019counterfactual} adopt a \ac{CDA} algorithm to mitigate social bias by swapping bias attribute words (e.g., he/she) in training dataset.
\citet{choi2022c2l} modify \ac{CDA} by masking the terms casual to label to force the model to learn label-invariant features.

In in-processing, \citet{guo2022auto,li2023prompt,yang2023adept} propose a two-stage adversarial method for debiasing.
They first train a continuous prompt to enlarge the bias of utterance pairs and then force the \acp{LLM} to minimize the difference of utterance pairs using the prompt.

In post-processing, \citet{schick2021self} propose a decoding algorithm that reduces the probability of a model producing biased text.
They use a textual description of the undesired behaviors for prompting.
\citet{meade2023using} propose an \ac{ICL} strategy which leverages non-biased demonstrations to guide the generation for safety.

However, existing methods for mitigating social bias either require the external bias knowledge for data augmentation and training adversarial prompts or require a non-biased dataset for building demonstrations, which are lacking for position debiasing and unpractical in application.
Differently, we propose a self-supervised framework for \acp{LLM} on mitigating position bias, without relying on any external bias knowledge or non-biased samples, which is general, simple but effective.

\section{Conclusion}
In this paper, we have proposed a self-supervised debiasing framework \model~for \acp{LLM}.
It adopts a multi-objective optimization module to mitigate position bias for \acp{LLM}, where unsupervised responses for the debiasing objective are of low position bias.
These responses are pruned by a proposed \ac{MSA} module for aligning the task and debiasing objectives.
Extensive experiments on five tasks and eight benchmark datasets show that \model~outperforms existing baselines on non-biased samples while retaining performance or sacrificing little performance on biased samples.
It demonstrates that leveraging unsupervised responses is a practicable solution to mitigate position bias for generative \acp{LLM}.

\section*{Limitations}
This work has the following limitations.
First, \model~needs a pre-trained \ac{LLM} to generate unsupervised responses with low bias, where these responses are still biased.
Second, the final performance of \model~depends on the quality of unsupervised responses, which are still noisy after being aligned by \ac{MSA}.
In future work, we plan to address these issues by investigating non-biased models from other domains and model-based strategies to align unsupervised responses.

\section*{Ethical Considerations}
We realize that there are risks in developing generative \acp{LLM}, so it is necessary to pay attention to the ethical issues of \acp{LLM}.
We use publicly available pre-trained \acp{LLM}, i.e., FlanT5-base, FlanT5-large, FlanT5-xlarge, and publicly available datasets in the academic community, i.e., CANARD, CoQAR, CNN/DM, Newsroom, Doc2dial, Mutual, SNLI, QNLI, to conduct experiments.
All models and datasets are carefully processed by their publishers to ensure that there are no ethical problems.

\section*{Acknowledgments}
We thank the reviewers for their valuable feedback.
This work was supported by the National Key R\&D Program of China with grant No.2022YFC3303004, the Natural Science Foundation of China (62372275, 62272274, 62202271, T2293773, 62102234, 62072279), the Natural Science Foundation of Shandong Province (ZR2021QF129).

\bibliography{acl.bib}

\clearpage

\appendix
\section{Experimental Details}
\subsection{Datasets} \label{sec: app_datasets}
We conduct experiments on eight widely used benchmark datasets: CANARD, CoQAR, CNN/DM, Newsroom, Doc2dial, Mutual, SNLI and QNLI, covering five \ac{NLP} tasks: \ac{CQA}, \ac{CQG}, summarization, \ac{KGC} and \ac{NLI}.
\begin{itemize}[leftmargin=*,nosep]
\item \textbf{CANARD}~\citep{CANARD} is a benchmark dataset for \ac{CQA} and \ac{CQG}.
It is built based on QuAC~\citep{QuAC} and consists of 40k questions with different context lengths, where answers are selected spans from a given section in a Wikipedia article.

\item \textbf{CoQAR}~\citep{brabant2022coqar} is a large-scale dataset for \ac{CQA} and \ac{CQG}.
It annotates 53k questions based on CoQA~\citep{reddy2019coqa}, where the documents are from seven diverse domains.

\item \textbf{CNN/DM}~\citep{nallapati2016abstractive} is a well-known summarization dataset, which consists of 313k articles from CNN and Daily Mail. 
The summary is written by human experts and shown as bullet points. 
We use the non-anonymized version~\cite {see2017get}. 

\item \textbf{Newsroom}~\citep{grusky2018newsroom} is a large-scale summarization dataset which contains 1.3 million articles and expert-written summaries with high diversity.

\item \textbf{Doc2dial}~\citep{feng2020doc2dial} is a document-grounded dialogue dataset with 4,800 annotated conversations and an average of 14 turns. 
Compared to the prior document-grounded dialogue datasets, this dataset covers a variety of dialogue scenes in information-seeking conversations.

\item \textbf{Mutual}~\citep{cui2020mutual} is a multi-turn reasoning dialogue dataset, consisting of 8,860 manually annotated dialogues based on Chinese student English listening comprehension exams.
It is challenging since it requires a model to handle various reasoning problems.

\item \textbf{SNLI}~\citep{bowman2015large} is a large-scale natural language inference benchmark with 570k utterance pairs. 
Each pair is manually labeled as entailment, neutral, or contradiction with several annotators.

\item \textbf{QNLI}~\citep{wang2018glue} is a natural language inference dataset derived from the Stanford Question Answering Dataset v1.1.
An utterance is extracted from the passage and paired with the question.
Each pair is then manually labeled according to whether the utterance contains the answer to the question.

\end{itemize}

\subsection{Bias Types and Dataset Splitting} \label{sec: app_dataset_split}
In this work, we focus on mitigating three types of widely addressed position bias: lead bias~\citep{kedzie2018content}, relative position bias~\citep{shinoda2022look} and lexical bias~\citep{gururangan2018annotation}.
\begin{itemize}[leftmargin=*, itemsep=0pt, topsep=5pt]
    \item Lead bias in summarization is a phenomenon that the generated summary is highly correlated to utterances appearing at the beginning of the document~\citep{kedzie2018content}.
    \item Relative position bias in QA is a phenomenon that a QA model tends to degrade the performance on samples where answers are located in relative positions unseen during training~\citep{shinoda2022look}.
    The relative position is defined as the relative position of grounded utterances between the last turn answer and the current turn answer.
    \item Lexical bias is the phenomenon that deep learning models achieve high accuracy by exploiting trigger words or word overlapping~\citep{gururangan2018annotation,poliak2018hypothesis}.
    Note that lexical bias is a type of position bias in generative models.
    For example, trigger words in hypothesis in generative models are regarded as trigger words behind `Hypothesis: ', which is a positional clue for prediction. 
\end{itemize}

Following previous works~\citep{ko2020look,shinoda2022look}, we split the test dataset into biased dataset and non-biased dataset for simulation depending on the bias type in each dataset.
In \ac{CQG} and \ac{CQA} datasets (CANARD and CoQAR), we select the samples with relative position equaling 0 or 1 into the biased dataset and the left samples into the non-biased dataset.
In \ac{KGC} and summarization datasets (Doc2dial, Mutual, CNN/DM and Newsroom), we filter samples where the reference response is highly correlated to the beginning utterance of the given document into the biased dataset and the left samples into the non-biased dataset.
For \ac{NLI} datasets (SNLI and QNLI), samples with specific words are filtered into the biased dataset and other samples are filtered into the non-biased dataset.
The dataset statistics are shown in Table~\ref{tab: dataset_stat}.

\begin{table}[t]
\centering
\caption{Dataset statistics after splitting.}
\label{tab: dataset_stat} 
\newcolumntype{M}[1]{>{\centering\arraybackslash}m{#1}}

\begin{tabular}{l M{20pt} M{20pt} M{25pt} M{25pt}}
\toprule
\multirow{2}{*}{Dataset} & \multicolumn{3}{c}{Biased} & Non-biased \\
\cmidrule{2-4}
 & \#Train & \#Dev & \#Test & \#Test \\
\midrule
CANARD & 500 & 250 & 3,460 & 2,440 \\ 
CoQAR & 500 & 250 & 3,222 & 4,873 \\
CNN/DM & 500 & 250 & 1,421 & 5,000 \\
Newsroom & 500 & 200 & 5,000 & 5,000 \\
Doc2dial & 500 & 250 & 2,000 & 5,000 \\
Mutual & 500 & 250 & 2,000 & 5,000 \\
SNLI & 500 & 250 & 2,000 & 5,000 \\
QNLI & 500 & 250 & 2,000 & 5,000 \\
\bottomrule
\end{tabular}

\end{table}

\newpage

\section{Analysis}
\subsection{Case Study} \label{sec: case_study}

\begin{table*}[t]

\setlength\tabcolsep{5pt}
\caption{An example of generated answers on CANARD. U1, U4, and U5 are the 1st, 4th, 5th utterances in the document, T1 and T2 are the first and second turn utterances of the dialogue. 
`Target' is the target answer of this example, `FT' and `\acs{ZOE}' are the generated answers of FT and \acs{ZOE}.
`Position' is the position of the grounded utterance of the answer in the document and `Relative position' is the distance of grounded utterances between the current answer and the last turn answer.}
\label{tab: case_study} 
\centering
\begin{tabular}{m{1.2cm}<{\centering} p{0.68\textwidth} c}
\toprule
ID & \makecell[c]{Document} &  \\
\cmidrule(lr){1-3}
U1 & \multicolumn{2}{p{0.865\textwidth}}{\textcolor{blue}{Gaston had worked with players at an individual level} as a hitting instructor ...}\\
\rowcolor{gray!10}
U4 & \multicolumn{2}{p{0.865\textwidth}}{\textcolor{red}{In the six games the Blue Jays played in those places during World Series play} ... } \\
U5 & \multicolumn{2}{p{0.865\textwidth}}{... and \textcolor{orange}{Gaston was the first ever African-American manager to win a World Series.}} \\
\midrule 
ID & \makecell[c]{Context} & Position \\
\midrule
\multirow{3}{*}{T1} & Question: What happens in the series?  \newline
Answer: \textcolor{red}{In the six games the Blue Jays played in those places during World Series play}, the Jays went 4-2 ... & \multirow{3}{*}{U4} \\
\rowcolor{gray!10}
T2 & Question: What else did Cito do? &  \\
\midrule
Model & \makecell[c]{Answer} & Relative position \\
\midrule
Target & \textcolor{blue}{Gaston had worked with players at an individual level} ... & -3 (U1) \\
\rowcolor{gray!10}
\multirow{2}{*}{FT} & ... \textcolor{orange}{Gaston was the first ever African-American manager to win a World Series.} & \multirow{2}{*}{\ 1 (U5)} \\
\acs{ZOE} & \textcolor{blue}{Gaston had worked with players at an individual level} ... & -3 (U1) \\
\bottomrule
\end{tabular}
\end{table*}

\begin{table*}[t]

\setlength\tabcolsep{5pt}
\caption{An example of unsupervised questions on CANARD. U1--U6 are the first six utterances in the document and T1 is the first turn utterance of the dialogue. 
`Target' is the target question of this example, `BASE-1' and `BASE-2' are the questions generated by pre-trained LLM.
`Position' is the position of the grounded utterance of the question in the document and `Relative position' is the distance of grounded utterances between the current question and the last turn question.}
\label{tab: case_study2} 
\centering
\begin{tabular}{m{1.5cm}<{\centering} p{0.68\textwidth} c}
\toprule
ID & \makecell[c]{Document} &  \\
\cmidrule(lr){1-3}
U1 & \multicolumn{2}{p{0.865\textwidth}}{Gautam Gambhir, \textcolor{orange}{born 14 October 1981}, is an Indian cricketer, ...} \\
\rowcolor{gray!10}
\multirow{2}{*}{U2} & \multicolumn{2}{p{0.865\textwidth}}{Gambhir was picked up by the Delhi Daredevils franchise in the first player auction of the \textcolor{orange}{Indian Premier League} for a price of US\$725,000 a year.}\\
U3 & \multicolumn{2}{p{0.865\textwidth}}{He became the second highest run-scorer of the inaugural season with ...} \\
\rowcolor{gray!10}
U4 & \multicolumn{2}{p{0.865\textwidth}}{He was \textcolor{blue}{promoted to the post of Captain of the Delhi Daredevils} for IPL Season 2010.} \\
\multirow{2}{*}{U5} & \multicolumn{2}{p{0.865\textwidth}}{At the end of the tournament he became the only player from Delhi Daredevils \textcolor{red}{to score more than 1000 runs in the IPL.}} \\
\rowcolor{gray!10}
U6 & \multicolumn{2}{p{0.865\textwidth}}{In the 2011 IPL player auction, ...} \\
\midrule 
ID & \makecell[c]{Context} & Position \\
\midrule
\multirow{3}{*}{T1} & Question: Who was Gautam Gambhir in Indian premier league?  \newline
Answer: He was \textcolor{blue}{promoted to the post of Captain of the Delhi Daredevils for IPL Season 2010.} & \multirow{3}{*}{U4} \\
\midrule
Source & \makecell[c]{Question} & Relative position \\
\midrule
Target & What did Gautam Gambhir \textcolor{red}{do as captain}? & \ 1 (U5) \\
\rowcolor{gray!10}
BASE-1 & Where was Gautam Gambhir \textcolor{orange}{born}? & -3 (U1) \\
BASE-2 & Was Gautam Gambhir in \textcolor{orange}{Indian premier league}? & -2 (U2) \\
\bottomrule
\end{tabular}

\end{table*}

To investigate the reason for the effectiveness of \model, we present an example of generated responses in \ac{CQA} in Table~\ref{tab: case_study} and an example of unsupervised responses in \ac{CQG} in Table~\ref{tab: case_study2}.

In \ac{CQA} example, \model~can generate an answer with lower bias than FT.
As shown in Table~\ref{tab: case_study}, FT generates an answer from U5, adjacent to U4, the utterance containing the last turn answer.
In contrast, \acs{ZOE} generates an answer from U1, which is far from U4.
After fine-tuning with unsupervised responses from various positions, \acs{ZOE} cannot easily fall into the trap of finding answers from neighboring utterances of the one containing the last turn answer.

In \ac{CQG} example in Table~\ref{tab: case_study2}, \model~uses unsupervised responses which have lower bias than the target response.
As we can see, target question is based on the utterance with relative position 1, while unsupervised questions generated from BASE are based on utterances varying in the document.
Fine-tuning with unsupervised responses generated from pre-trained \acp{LLM} encourages the model to generate questions based on utterances in various document positions, thus mitigating position bias.

\clearpage

\section{Overall Analysis} \label{sec: overall_analysis}

\hspace{-8pt} 
\begin{minipage}[t]{\textwidth}
\centering
\captionof{table}{\model~Performance (\%) using various unsupervised responses on language understanding tasks. Boldface indicates the best results in terms of the corresponding dataset.}
\label{tab: appendix_lu_task} 

\setlength{\tabcolsep}{4pt}
\begin{tabular}{l c c c c c c c c c}
\toprule
\multirow{3}{*}{Method}  & \multicolumn{4}{c}{NLI~(\%)} & \multicolumn{4}{c}{CQA~(\%)} \\
 & \multicolumn{2}{c}{SNLI} & \multicolumn{2}{c}{QNLI} & \multicolumn{2}{c}{CoQAR} & \multicolumn{2}{c}{CANARD} \\ 
\cmidrule(lr){2-3}\cmidrule(lr){4-5}\cmidrule(lr){6-7}\cmidrule(lr){8-9}
 & Biased & Non-Biased & Biased & Non-Biased & Biased & Non-Biased & Biased & Non-Biased \\
 \midrule
\acs{ZOE} & \textbf{92.0} & \textbf{88.4} & 94.3 & 90.9 & \textbf{66.3} & \textbf{53.7} & 65.7 & 21.9 \\
- w/o \acs{MSA} & 91.0 & 87.4 & 92.7 & 88.7 & \textbf{66.3} & \textbf{53.7} & 65.7 & 21.9 \\
- w/ T5-base & 90.5 & 87.6 & \textbf{94.4} & 90.6 & 64.3 & 52.3 & 61.6 & 19.9 \\
- w/ T5-xlarge & 91.2 & 87.7 & 93.2 & \textbf{91.3} & 65.1 & 53.6 & 67.1 & \textbf{22.4} \\
FT & \textbf{92.0} & 88.0 & 94.3 & 89.8 & 64.6 & 51.9 & \textbf{67.5} & 20.8 \\
\bottomrule
\end{tabular}

\end{minipage}

\vspace{\baselineskip}

\hspace{-18pt} 
\begin{minipage}[t]{\textwidth}
\centering
\captionof{table}{\model~Performance (\%) using various unsupervised responses on language creation tasks.
Boldface indicates the best results in terms of the corresponding dataset.}
\label{tab: appendix_lc_task} 

\setlength{\tabcolsep}{4pt}
\begin{tabular}{l c c c c c c c c c}
\toprule
\multirow{3}{*}{Method}  & \multicolumn{4}{c}{CQG~(\%)} & \multicolumn{4}{c}{KGC~(\%)} \\
 & \multicolumn{2}{c}{CoQAR} & \multicolumn{2}{c}{CANARD} & \multicolumn{2}{c}{Doc2dial} & \multicolumn{2}{c}{Mutual} \\ 
\cmidrule(lr){2-3}\cmidrule(lr){4-5}\cmidrule(lr){6-7}\cmidrule(lr){8-9}
 & Biased & Non-Biased & Biased & Non-Biased & Biased & Non-Biased & Biased & Non-Biased \\
 \midrule
\acs{ZOE} & \textbf{26.7} & \textbf{18.8} & 25.9 & \textbf{22.4} & 40.6 & 38.3 & 94.0 & \textbf{53.0}     \\ 
- w/o \acs{MSA} & 25.8 & 18.0 & \textbf{26.1} & 21.5 & 32.6 & 34.1 & 91.0 & 36.7     \\ 
- w/ T5-base & 26.3 & 18.3 & \textbf{26.1} & 22.0 & 42.8 & 36.3 & \textbf{94.3} & 45.1     \\ 
- w/ T5-xlarge & 26.6 & 17.9 & 25.7 & 22.0 & \textbf{44.1} & \textbf{39.1} & \textbf{94.3} & 52.7     \\ 
FT & \textbf{26.7} & 17.2 & 26.0 & 21.6 & 38.7 & 36.0 & 93.9 & 38.9 \\
\bottomrule
\end{tabular}
\end{minipage}

\vspace{\baselineskip}

\hspace{-18pt} 
\begin{minipage}[t]{0.48\textwidth}
\centering
\captionof{table}{\model~Performance (\%) using various unsupervised responses on language compression tasks.
`N-Biased' denotes performance on the non-biased datasets.
Boldface indicates the best results in terms of the corresponding dataset.}
\label{tab: appendix_lcomp_task} 

\setlength{\tabcolsep}{2pt}
\begin{tabular}{l c c c c c}
\toprule
\multirow{3}{*}{Method} & \multicolumn{4}{c}{Summarization~(\%)} \\ 
 & \multicolumn{2}{c}{CNN/DM} & \multicolumn{2}{c}{Newsroom} \\ 
\cmidrule(lr){2-3}\cmidrule(lr){4-5}
 & Biased & N-Biased & Biased & N-Biased \\
 \midrule
\acs{ZOE} & 27.1 & \textbf{17.3} & 50.9 & \textbf{21.3} \\
- w/o \acs{MSA} & 27.1 & 14.8 & 48.0 & 20.2     \\
- w/ T5-base & 26.8 & 14.2 & 47.7 & 20.3     \\
- w/ T5-xlarge & \textbf{28.0} & 14.6 & 48.4 & 20.2     \\ 
FT & 26.9 & 16.8 & \textbf{51.1} & 19.8 \\
\bottomrule
\end{tabular}

\end{minipage}

\subsection{Analyses of Unsupervised Responses}
Table~\ref{tab: appendix_lu_task}-\ref{tab: appendix_lcomp_task} demonstrate \model~performance using different qualities of unsupervised responses.
As we can see, lower response quality leads to worse performance.
Besides, using various sources of unsupervised responses always degrades model performance on non-biased datasets.
\newpage

\vspace*{25\baselineskip}

\subsection{Analyses of Objective Weighting}
\begin{figure}[th]
    \centering
    
    \subfloat[CANARD~(CQG)]{\vspace{-5pt}\includegraphics[width=0.23\textwidth]{alpha_canard_qg.pdf}}
    \hfill
    \subfloat[CoQAR~(CQG)]{\vspace{-5pt}\includegraphics[width=0.23\textwidth]{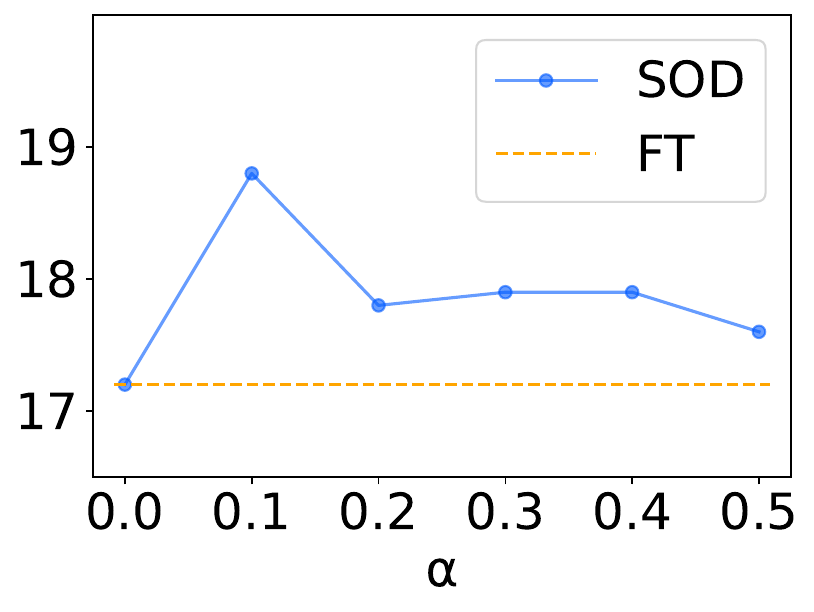}}
    \newline
    \subfloat[CANARD~(CQA)]{\vspace{-5pt}\includegraphics[width=0.23\textwidth]{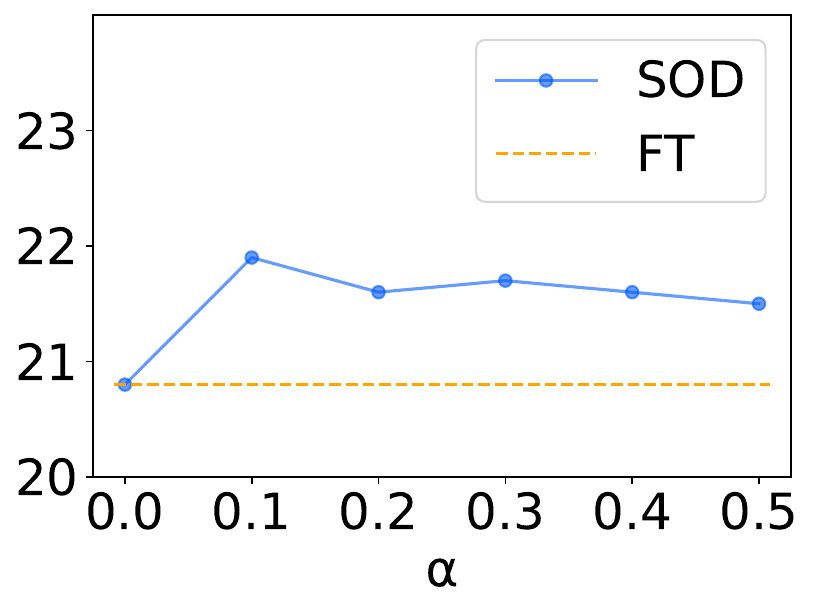}}
    \hfill
    \subfloat[CoQAR~(CQA)]{\vspace{-5pt}\includegraphics[width=0.23\textwidth]{alpha_coqa_qa.pdf}}
    \newline
    \subfloat[Newsroom]{\vspace{-5pt}\includegraphics[width=0.23\textwidth]{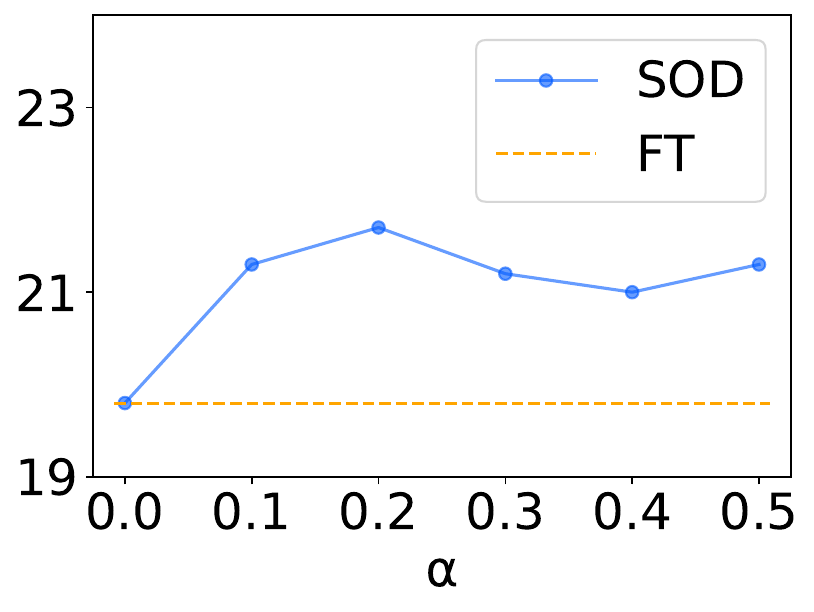}}
    \hfill
    \subfloat[CNN/DM]{\vspace{-5pt}\includegraphics[width=0.23\textwidth]{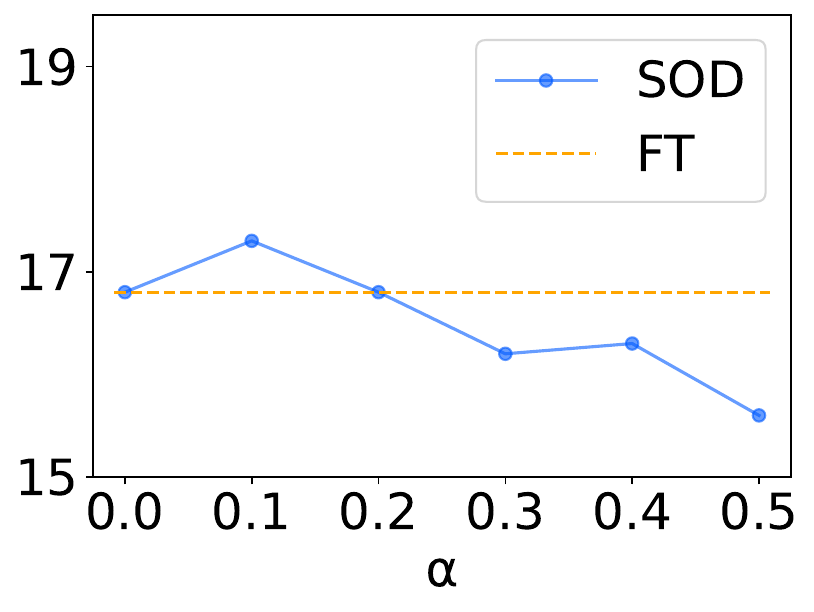}}
    \newline
    \subfloat[Doc2dial]{\vspace{-5pt}\includegraphics[width=0.23\textwidth]{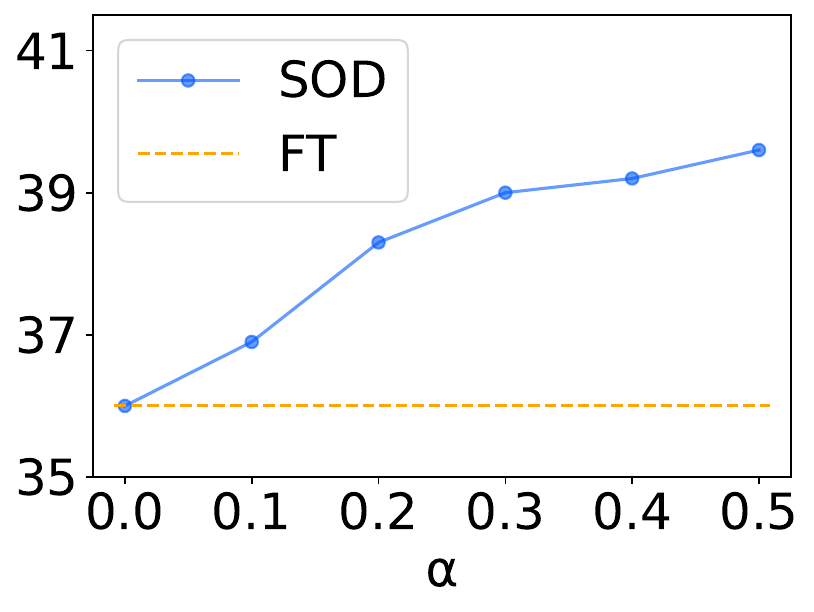}}
    \hfill
    \subfloat[Mutual]{\vspace{-5pt}\includegraphics[width=0.23\textwidth]{alpha_mutual.pdf}}
    \newline
    \subfloat[SNLI]{\vspace{-5pt}\includegraphics[width=0.23\textwidth]{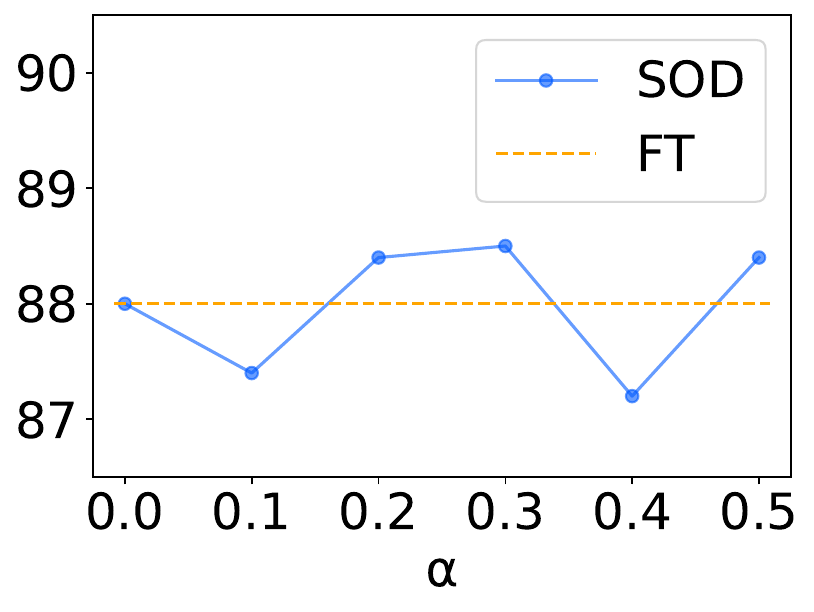}}
    \hfill
    \subfloat[QNLI]{\vspace{-5pt}\includegraphics[width=0.23\textwidth]{alpha_qnli.pdf}}
    \caption{Performance over each $\alpha$ on all datasets. 
    The x-axis denotes the value of $\alpha$ and the y-axis denotes the ROUGE-L score on non-biased datasets.}
    \label{fig:appendix_range_alpha}

\end{figure}

Fig.~\ref{fig:appendix_range_alpha} provides the overall performance over different $\alpha$.
In most cases, \model~outperforms the fine-tuned LLM, i.e., FT.

\subsection{Analyses on Training Samples}
\begin{figure}[H]
    \centering
    \subfloat[CANARD~(CQG)]{\vspace{-5pt}\includegraphics[width=0.23\textwidth]{num_canard_qg.pdf}}
    \hfill
    \subfloat[CoQAR~(CQG)]{\vspace{-5pt}\includegraphics[width=0.23\textwidth]{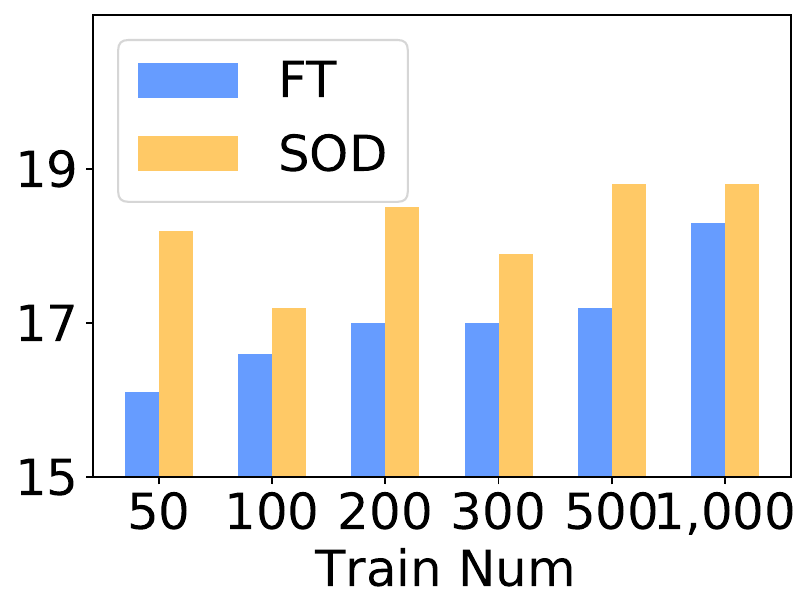}}
    \newline
    \subfloat[CANARD~(CQA)]{\vspace{-5pt}\includegraphics[width=0.23\textwidth]{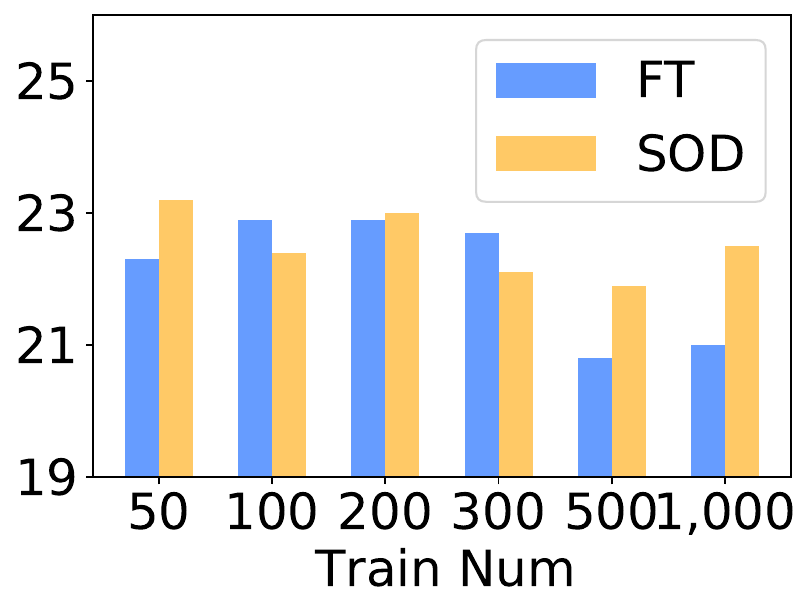}}
    \hfill
    \subfloat[CoQAR~(CQA)]{\vspace{-5pt}\includegraphics[width=0.23\textwidth]{num_coqa_qa.pdf}}
    \newline
    \subfloat[Newsroom]{\vspace{-5pt}\includegraphics[width=0.23\textwidth]{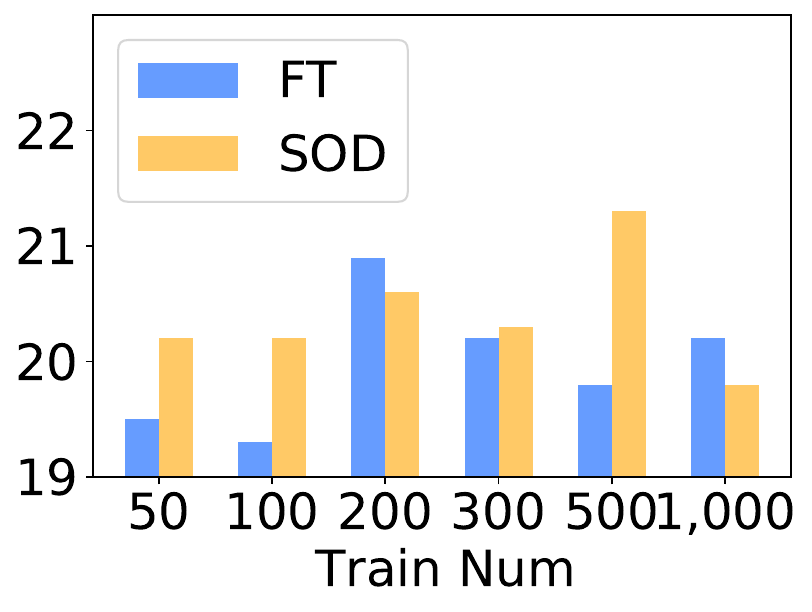}}
    \hfill
    \subfloat[CNN/DM]{\vspace{-5pt}\includegraphics[width=0.23\textwidth]{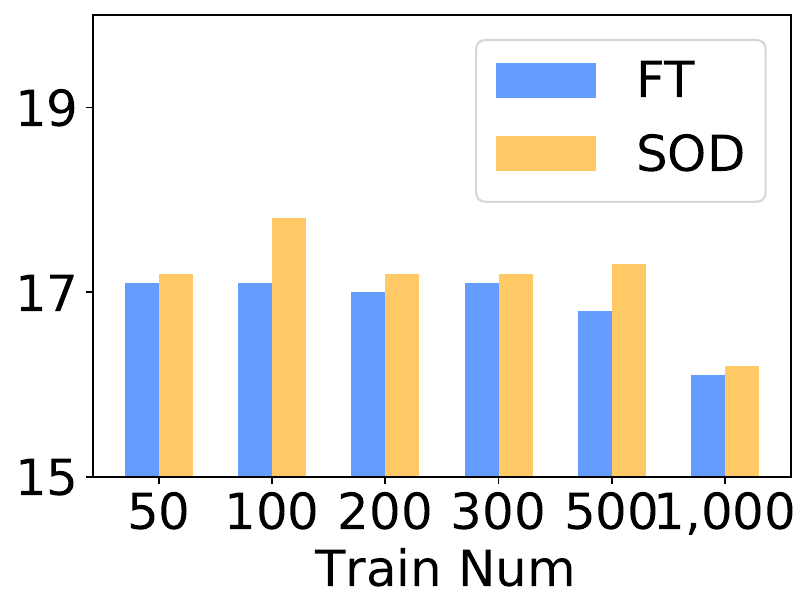}}
    \newline
    \subfloat[Doc2dial]{\vspace{-5pt}\includegraphics[width=0.23\textwidth]{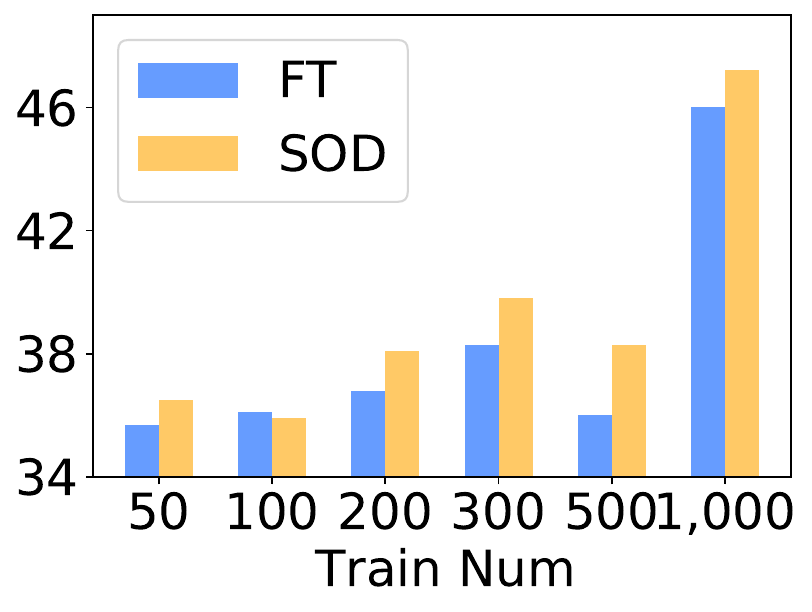}}
    \hfill
    \subfloat[Mutual]{\vspace{-5pt}\includegraphics[width=0.23\textwidth]{num_mutual.pdf}}
    \newline
    \subfloat[SNLI]{\vspace{-5pt}\includegraphics[width=0.23\textwidth]{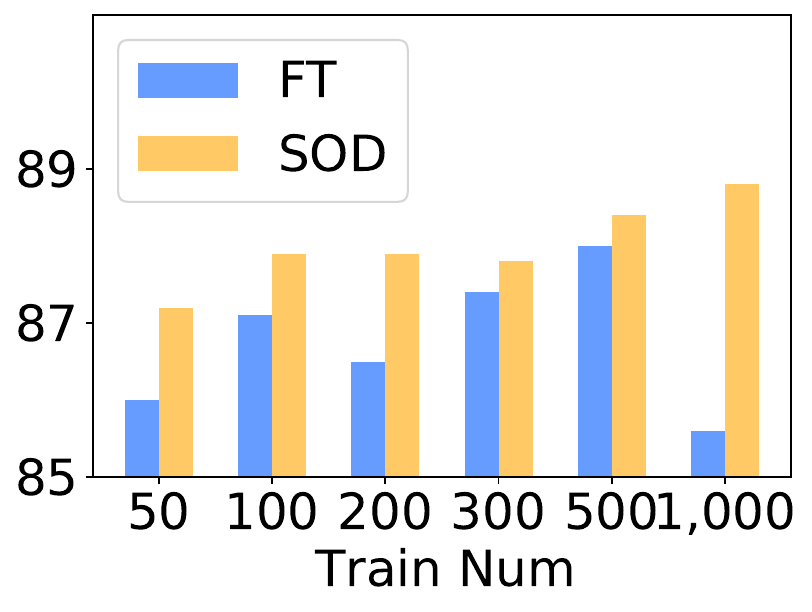}}
    \hfill
    \subfloat[QNLI]{\vspace{-5pt}\includegraphics[width=0.23\textwidth]{num_qnli.pdf}}
    \caption{Performance over different numbers of training samples on all datasets. 
    The x-axis denotes the number of training samples and the y-axis denotes the ROUGE-L score on non-biased datasets.}
    \label{fig:appendix_range_num}

\end{figure}

Fig.~\ref{fig:appendix_range_num} provides the overall performance on different numbers of training samples.
In most cases, \model~outperforms the fine-tuned LLM, i.e., FT.

\subsection{Cases} \label{sec: cases}
We also present cases for \ac{CQA}, \ac{CQG}, summarization and \ac{KGC} tasks in Table~\ref{tab: app_case_cqa_coqar}--\ref{tab: app_case_kgc_mutual}.
As we can see, FT always finds relevant information from utterances near the grounded utterances of the last turn utterance for \ac{CQA} and \ac{CQG} tasks or from the lead utterance for summarization and \ac{KGC} tasks.
While \model~can find relevant knowledge from any utterances in the document.
Note that on the Mutual dataset, where each sample has four candidate responses as the document, FT always fails to select the relevant utterance from the last two responses.

\clearpage

\clearpage

\begin{table*}
\setlength\tabcolsep{5pt}
\caption{An example of \ac{CQA} on CoQAR. U1--U4 are the first four utterances in the document, T1 and T2 are the 1st and 2nd turn utterances of the dialogue. 
`Target' is the target answer of the example, `FT' and `\acs{ZOE}' are the generated answers from FT and \acs{ZOE}.
`Position' is the position of the grounded utterance of the answer in the document and `Relative position' is the distance of grounded utterances between the current answer and the last turn answer.}
\label{tab: app_case_cqa_coqar} 
\centering
\begin{tabular}{m{1.2cm}<{\centering} p{0.68\textwidth} c}
\toprule
ID & \makecell[c]{Document} &  \\
\cmidrule(lr){1-3}
U1 & \multicolumn{2}{p{0.865\textwidth}}{This final war was to give thousands of colonists, including ..., \textcolor{red}{military experience} ...}\\
\rowcolor{gray!10}
\multirow{2}{*}{U2} & \multicolumn{2}{p{0.865\textwidth}}{By far the largest military action in which the United States engaged during this era was \textcolor{red}{the War of 1812}.} \\
\multirow{2}{*}{U3} & \multicolumn{2}{p{0.865\textwidth}}{With \textcolor{orange}{Britain} locked in a major war with Napoleon's France, its policy was to block American shipments to France} \\
\rowcolor{gray!10}
U4 & \multicolumn{2}{p{0.865\textwidth}}{The United States sought to remain \textcolor{blue}{neutral} while pursuing overseas trade.} \\
\midrule 
ID & \makecell[c]{Context} & Position \\
\midrule
\multirow{2}{*}{T1} & Question: What did this give the colonists?  \newline
Answer: \textcolor{red}{Military experience}. & \multirow{2}{*}{U1} \\
\rowcolor{gray!10}
T2 & Question: Whose side was the US on at first in \textcolor{red}{the war of 1812}? & \\
\midrule
Model & \makecell[c]{Answer} & Relative position \\
\midrule
Target & At first \textcolor{blue}{neutral} & 3 (U4) \\
\rowcolor{gray!10}
FT & \textcolor{orange}{Britain.} & 2 (U3) \\
\acs{ZOE} & \textcolor{blue}{Neutral} & 3 (U4) \\
\bottomrule
\end{tabular}
\end{table*}

\begin{table*}
\setlength\tabcolsep{5pt}
\caption{An example of \ac{CQG} on CoQA. U1--U5 are the first five utterances in the document and T5 is the \textnormal{5th} turn utterance of the dialogue. 
`Target' is the target question of the example and `FT' and `\acs{ZOE}' are the generated questions from FT and \acs{ZOE}.
`Position' is the position of the grounded utterance of the question in the document and `Relative position' is the distance of grounded utterances between the current question and the last turn question.}
\label{tab: app_case_cqg_coqa} 
\centering
\begin{tabular}{m{1.2cm}<{\centering} p{0.68\textwidth} c}
\toprule
ID & \makecell[c]{Document} &  \\
\cmidrule(lr){1-3}
\multirow{2}{*}{U1} & \multicolumn{2}{p{0.865\textwidth}}{John's Metropolitan Area is the \textcolor{blue}{second largest Census Metropolitan Area (CMA) in Atlantic Canada} ...}\\
\rowcolor{gray!10}
\multirow{2}{*}{U3} & \multicolumn{2}{p{0.865\textwidth}}{Its name has been attributed to the feast day of John the Baptist, when John Cabot was believed to have \textcolor{cyan}{sailed into the harbor in 1497}.} \\
\multirow{2}{*}{U4} & \multicolumn{2}{p{0.865\textwidth}}{St. John's is one of the oldest settlements in \textcolor{red}{North America}, with year-round \textcolor{orange}{settlement beginning sometime after 1630} and seasonal habitation long before that.} \\
\rowcolor{gray!10}
U5 & \multicolumn{2}{p{0.865\textwidth}}{It is not, however, the oldest surviving English settlement in North America or Canada ...} \\
\midrule 
ID & \makecell[c]{Context} & Position \\
\midrule
\multirow{2}{*}{T5} & Question: In what continent is it located? \newline Answer: \textcolor{red}{North America}. & \multirow{2}{*}{U4} \\
\midrule
Model & \makecell[c]{Question} & Relative position \\
\midrule
Target & What year did John Cabot \textcolor{cyan}{arrive in the harbor?} & -1 (U3) \\
\rowcolor{gray!10}
FT & When was \textcolor{orange}{St. John's founded?} & \ 0 (U4) \\
\acs{ZOE} & What is the \textcolor{blue}{second largest CMA in Atlantic Canada}? & -3 (U1) \\
\bottomrule
\end{tabular}
\end{table*}

\begin{table*}
\setlength\tabcolsep{5pt}
\caption{An example of \ac{CQG} on CANARD. U1--U5 are the first five utterances in the document and T1 is the \textnormal{1st} turn utterance of the dialogue. 
`Target' is the target question of the example and `FT' and `\acs{ZOE}' are the generated questions from FT and \acs{ZOE}.
`Position' is the position of the grounded utterance of the question in the document and `Relative position' is the distance of grounded utterances between the current question and the last turn question.}
\label{tab: app_case_cqg_canard} 
\centering
\begin{tabular}{m{1.2cm}<{\centering} p{0.68\textwidth} c}
\toprule
ID & \makecell[c]{Document} &  \\
\cmidrule(lr){1-3}
\multirow{2}{*}{U1} & \multicolumn{2}{p{0.865\textwidth}}{In May 2009, production company CinemaNX announced that it would distribute Me and Orson Welles itself ... }\\
\rowcolor{gray!10}
\multirow{2}{*}{U2} & \multicolumn{2}{p{0.865\textwidth}}{It opened the New Orleans Film Festival on October 9, 2009; and it was screened at the St. Louis International Film Festival in November 2009.} \\
\multirow{2}{*}{U3} & \multicolumn{2}{p{0.865\textwidth}}{The film was \textcolor{red}{released in the US on November 25, 2009}, and \textcolor{orange}{in the UK on December 4, 2009}.} \\
\rowcolor{gray!10}
\multirow{3}{*}{U4} & \multicolumn{2}{p{0.865\textwidth}}{IndieWIRE reported, ``The do-it-yourself release of Richard Linklater's Me and Orson Welles bluegot off to a very nice start, averaging \$15,910 from its four theaters, the highest PTA of all debuting films.''} \\
\multirow{2}{*}{U5} & \multicolumn{2}{p{0.865\textwidth}}{While Orson Welles is \textcolor{blue}{one the first examples of such a high-profile film going to the DIY route}, if it proves successful, it's going to be done a lot more in the future.} \\
\midrule 
ID & \makecell[c]{Context} & Position \\
\midrule
\multirow{2}{*}{T1} & Question:  What month was Me and Orson Welles released in theaters? \newline Answer: The film was \textcolor{red}{released in the US on November 25, 2009}. & \multirow{2}{*}{U3} \\
\midrule
Model & \makecell[c]{Question} & Relative position \\
\midrule
Target & What were critics \textcolor{blue}{reviews} of Me and Orson Welles? & 2 (U5) \\
\rowcolor{gray!10}
FT & When was the film Me and Orson Welles \textcolor{orange}{released in the UK}? & 0 (U3) \\
\acs{ZOE} & What was the \textcolor{blue}{response} to Me and Orson Welles? & 2 (U5) \\
\bottomrule
\end{tabular}
\end{table*}

\begin{table*}
\setlength\tabcolsep{5pt}
\caption{An example of summarization on Newsroom. U1, U2, U7 and U8 are the 1st, 2nd, 7th and 8th utterances in the document. 
`Target' is the target summary of the document and `FT' and `\acs{ZOE}' are the generated summary from FT and \acs{ZOE}.
`Position' is the position of the utterance associated with the summary in the document.}
\label{tab: app_case_sum_news} 
\centering
\begin{tabular}{m{1.2cm}<{\centering} p{0.68\textwidth} c}
\toprule
ID & \makecell[c]{Document} &  \\
\cmidrule(lr){1-3}
U1 & \multicolumn{2}{p{0.8\textwidth}}{\textcolor{orange}{Joe Staley celebrates Colin Kaepernick's touchdown run in Super Bowl XLVII.}} \\
\rowcolor{gray!10}
\multirow{2}{*}{U2} & \multicolumn{2}{p{0.8\textwidth}}{\textcolor{red}{Cubs inspire Super Bowl memories for Staley Cubs fans were euphoric Wednesday night.}} \\
U7 & \multicolumn{2}{p{0.8\textwidth}}{Staley, 32, is the longest-tenured member of a team that's 6-17 since 2015.} \\
\rowcolor{gray!10}
\multirow{2}{*}{U8} & \multicolumn{2}{p{0.8\textwidth}}{Does he think the 49ers, who are nearly 22 years removed from their last title, \textcolor{blue}{can win a Super Bowl before he retires? `` Yeah,''} he said, ``I'll never give up, man.''} \\
\midrule
Model & \makecell[c]{Summary} & Position \\
\midrule
\multirow{2}{*}{Target} & \textcolor{red}{Cubs inspire Super Bowl memories for Staley Cubs fans were euphoric Wednesday night} ... & \multirow{2}{*}{U2} \\
\rowcolor{gray!10}
\multirow{2}{*}{FT} & \textcolor{orange}{Joe Staley celebrates Colin Kaepernick's touchdown run in Super Bowl XLVII.} Cubs inspire Super Bowl memories for Staley. & \multirow{2}{*}{U1} \\
\multirow{2}{*}{\acs{ZOE}} & \textcolor{blue}{Joe Staley says he wished the 49ers had won the Super Bowl after the 2012 season.} & \multirow{2}{*}{U8} \\
\bottomrule
\end{tabular}
\end{table*}

\begin{table*}
\setlength\tabcolsep{5pt}
\caption{An example of summarization on CNN/DM. U1--U3 are the first three utterances in the document. 
`Target' is the target summary of the document and `FT' and `\acs{ZOE}' are the generated summary from FT and \acs{ZOE}.
`Position' is the position of the utterance associated with the summary in the document.}
\label{tab: app_case_sum_cnndm} 
\centering
\begin{tabular}{m{1.2cm}<{\centering} p{0.68\textwidth} c}
\toprule
ID & \makecell[c]{Document} &  \\
\cmidrule(lr){1-3}
\multirow{2}{*}{U1} & \multicolumn{2}{p{0.8\textwidth}}{Negotiations between the United States and Libya that could result in \textcolor{orange}{compensation for past acts of state-sponsored terrorism by Libya} are under way.} \\
\rowcolor{gray!10}
\multirow{2}{*}{U2} & \multicolumn{2}{p{0.8\textwidth}}{The wreckage of Pan Am 103 in Lockerbie, Scotland; the bombing killed 270 people in 1989. U.S. and Libyan officials met Wednesday and Thursday, the official said.} \\
\multirow{2}{*}{U3} & \multicolumn{2}{p{0.8\textwidth}}{The nations hope to hammer out a deal in which \textcolor{blue}{Libya would ``resolve all outstanding claims in good faith''} and offer \textcolor{red}{``fair compensation'' to victims} and their families, he said.} \\
\midrule
Model & \makecell[c]{Summary} & Position \\
\midrule
\multirow{2}{*}{Target} & \textcolor{red}{The negotiations could result in compensation for past acts of state-sponsored terrorism by Libya}. & \multirow{2}{*}{U3} \\
\rowcolor{gray!10}
\multirow{2}{*}{FT} & The nations hope to hammer out a deal in which Libya \textcolor{orange}{would ``resolve all outstanding claims in good faith''}. & \multirow{2}{*}{U1} \\
\multirow{2}{*}{\acs{ZOE}} & Nations hoping for deal in which \textcolor{blue}{Libya would compensate terrorism victims}. & \multirow{2}{*}{U3} \\
\bottomrule
\end{tabular}
\end{table*}

\begin{table*}
\setlength\tabcolsep{5pt}
\caption{An example of \ac{KGC} on Doc2dial. U7--U10 represent the consecutive four utterances following the \textnormal{7th} utterance in the document. 
T1 and T2 are the \textnormal{1st} and \textnormal{2nd} turn utterances of the dialogue. 
`Target' is the target response of the example, `FT' and `\acs{ZOE}' are the generated responses from FT and \acs{ZOE}.
`Position' is the position of the grounded utterance of the response in the document and `Relative position' is the distance of grounded utterances between the current response and the last turn response.}
\label{tab: app_case_kgc_dial} 
\centering
\begin{tabular}{m{1.2cm}<{\centering} p{0.68\textwidth} c}
\toprule
ID & \makecell[c]{Document} &  \\
\cmidrule(lr){1-3}
U7 & \multicolumn{2}{p{0.8\textwidth}}{Should I Sign Up For Medical Insurance (Part B)?}\\
\rowcolor{gray!10}
\multirow{2}{*}{U8} & \multicolumn{2}{p{0.865\textwidth}}{\textcolor{blue}{When you sign up for Medicare, you will be asked if you want to enroll in Medical Insurance Part B}.} \\
\multirow{3}{*}{U9} & \multicolumn{2}{p{0.865\textwidth}}{If you do not choose to enroll in Medicare Part B and then decide to do so later, your coverage maybe delayed and you may have to pay a higher monthly premium unless you qualify for a ``Special Enrollment Period'' or SEP.} \\
\rowcolor{gray!10}
U10 & \multicolumn{2}{p{0.865\textwidth}}{\textcolor{orange}{You may qualify for a SEP that will let you sign up for Part B} during ...} \\
\midrule 
ID & \makecell[c]{Context} & Position \\
\midrule
\multirow{4}{*}{T1} & Question: We don't need to get any supplementary medical insurance, isn't?  \newline
Answer: Yes, that's correct. \textcolor{orange}{You may qualify for a SEP that will let you sign up for Part B.} & \multirow{4}{*}{U10} \\
\rowcolor{gray!10}
T2 & Question: Do I sign up for medicare part B? & \\
\midrule
Model & \makecell[c]{Answer} & Relative position \\
\midrule
\multirow{2}{*}{Target} & \textcolor{blue}{When you sign up for Medicare you will be asked if you want to enroll in Medical insurance Part B.} & \multirow{2}{*}{-2 (U8)} \\
\rowcolor{gray!10}
\multirow{2}{*}{FT} & Yes, that's correct. \textcolor{orange}{You may qualify for a SEP that will let you sign up for Part B.} & \multirow{2}{*}{\ \ \ 0 (U10)} \\
\acs{ZOE} & \textcolor{blue}{Yes, you will be asked if you want to enroll in Medical insurance Part B.} & -2 (U8) \\
\bottomrule
\end{tabular}
\end{table*}

\clearpage

\begin{table*}
\vspace*{-30em}
\setlength\tabcolsep{5pt}
\caption{An example of \ac{KGC} on Mutual. U1--U4 represent the four candidate responses. 
T1 and T2 are the \textnormal{1st} and \textnormal{2nd} turn utterances of the dialogue. 
`Target' is the target response of the example, `FT' and `\acs{ZOE}' are the generated responses from FT and \acs{ZOE}.
`Position' is the position of the selected response.}
\label{tab: app_case_kgc_mutual} 
\centering
\begin{tabular}{m{1.2cm}<{\centering} p{0.68\textwidth} c}
\toprule
ID & \makecell[c]{Document} &  \\
\cmidrule(lr){1-3}
\multirow{2}{*}{U1} & \multicolumn{2}{p{0.8\textwidth}}{\textcolor{orange}{It does n't matter. you just joined a new team, and the manager said it's normal that you are not good at interpersonal skills.}}\\
\rowcolor{gray!10}
\multirow{2}{*}{U2} & \multicolumn{2}{p{0.8\textwidth}}{Although the manager said you are not good at interpersonal skills, you still evaluated others' performances.} \\
\multirow{2}{*}{U3} & \multicolumn{2}{p{0.8\textwidth}}{So you had your performance evaluation yesterday and were praised by the manager, right?} \\
\rowcolor{gray!10}
U4 & \multicolumn{2}{p{0.8\textwidth}}{\textcolor{blue}{yeah, you were praised by the manager, weren't you?}} \\
\midrule 
ID & \makecell[c]{Context} & Position \\
\midrule
\multirow{2}{*}{T1} & Female: you look happy.  \newline
Male: I am. I had my performance evaluation today. & \multirow{2}{*}{--} \\
\rowcolor{gray!10}
\multirow{3}{*}{T2} & Female: so it went well?  \newline
Male: yes, the manager said my interpersonal skills are great. I work well with others. & \multirow{3}{*}{--} \\
\midrule
Model & \makecell[c]{Female} & Position \\
\midrule
Target & \textcolor{blue}{Yeah, you were praised by the manager, weren't you?} & U4 \\
\rowcolor{gray!10}
\multirow{2}{*}{FT} & \textcolor{orange}{It does n't matter. you just joined a new team, and the manager said it's normal that you are not good at interpersonal skills.} & \multirow{2}{*}{U1} \\
\acs{ZOE} & \textcolor{blue}{Yeah, you were praised by the manager, weren't you?} & U4 \\
\bottomrule
\end{tabular}
\end{table*}

\end{document}